\documentclass[10pt,twocolumn,letterpaper]{article}
\pdfoutput=1

\usepackage{iccv}
\usepackage{times}
\usepackage{epsfig}
\usepackage{graphicx}
\usepackage{amsmath}
\usepackage{amssymb}
\usepackage{amssymb}

\usepackage{booktabs}
\usepackage{multirow}
\usepackage{xcolor} 
\usepackage{inconsolata}
\usepackage{enumerate}
\usepackage[accsupp]{axessibility}
\usepackage{amsmath,amsfonts,relsize,graphicx,dsfont}
\usepackage{color}

\makeatletter
\newcommand*\smallE{\mathds{E}} 
\newcommand*\yo@bigE[2]{\text{#2\raisebox{-#1ex}{$\smallE$}}}
\DeclareRobustCommand*\bigE{\mathop{\mathchoice
  {\yo@bigE{0.5}{\larger\larger}} 
  {\yo@bigE{0.4}{\larger\larger}} 
  {\yo@bigE{0.3}{\larger\larger}} 
  {\text{\scalebox{1.0}{\raisebox{-0.5ex}{$\smallE$}}}} 
  }\displaylimits}
\makeatother

\newenvironment{packed_enumerate}
{\begin{enumerate}
    \vspace{-\topsep}
    \setlength{\itemsep}{1pt}
    \setlength{\parskip}{0pt}
    \setlength{\parsep}{0pt}
}{\end{enumerate}}

\newcommand{\methodname}[1]{\textit{Sparse Intermittent Rewrite Injection}\xspace}
\newcommand{\methodshort}[1]{\textsc{SIRI}\xspace}

\newcommand{\search}[1][1]{\textit{Search}\xspace}
\newcommand{\rewrite}[1][1]{\textit{Rewrite}\xspace}
\newcommand{\train}[1][1]{\textit{Train}\xspace}

\newcommand{\ablation}[1][1]{PLAD\xspace}
\newcommand{\pplusr}[1][1]{\textit{Naive Rewrite Integration}\xspace}
\newcommand{\pplusrshort}[1][1]{\textsc{NRI}\xspace}

\newcommand{\RWS}[1][1]{\textit{RWS}\xspace}
\newcommand{\RW}[1][1]{\textit{RW}\xspace}
\newcommand{\Obj}[1][1]{$\mathcal{O}$\xspace}

\newcommand{\doname}[1][1]{\textit{Parameter Optimization}\xspace}
\newcommand{\doshort}[1][1]{\textit{PO}\xspace}

\newcommand{\cpname}[1][1]{\textit{Code Pruning}\xspace}
\newcommand{\cpshort}[1][1]{\textit{CP}\xspace}

\newcommand{\cgname}[1][1]{\textit{Code Grafting}\xspace}
\newcommand{\cgshort}[1][1]{\textit{CG}\xspace}

\DeclareMathOperator*{\argmax}{arg\,max}

\newcommand{\rone}[1][1]{\textsc{\color{blue}R1}\xspace}
\newcommand{\rtwo}[1][1]{\textsc{\color{green}R2}\xspace}
\newcommand{\rthree}[1][1]{\textsc{\color{red}R3}\xspace}

\newcommand\blfootnote[1]{%
  \begingroup
  \renewcommand\thefootnote{}\footnote{#1}%
  \addtocounter{footnote}{-1}%
  \endgroup
}

\usepackage{array}

\RequirePackage{xr-hyper}
\usepackage[pagebackref=true,breaklinks=true,colorlinks,bookmarks=false]{hyperref}
\usepackage[capitalize]{cleveref}
\crefname{section}{Sec.}{Secs.}
\Crefname{section}{Section}{Sections}
\Crefname{table}{Table}{Tables}
\crefname{table}{Tab.}{Tabs.}

\iccvfinalcopy 

\def\iccvPaperID{3261} 

\ificcvfinal\fi

\begin{document}

\title{Improving Unsupervised Visual Program Inference with\\Code Rewriting Families}

\author{Aditya Ganeshan \qquad R. Kenny Jones \qquad Daniel Ritchie\\
Brown University\\
{\tt\small adityaganeshan@gmail.com} \\
}

\maketitle

\ificcvfinal\thispagestyle{empty}\fi
\begin{abstract}
Programs offer compactness and structure that makes them an attractive representation for visual data.
We explore how \emph{code rewriting} can be used to improve systems for inferring programs from visual data.
We first propose \methodname~ (\methodshort~), a framework for unsupervised bootstrapped learning.
\methodshort~ sparsely applies code rewrite operations over a dataset of training programs, injecting the improved programs back into the training set.
We design a family of rewriters for visual programming domains: parameter optimization, code pruning, and code grafting.
For three shape programming languages in 2D and 3D, we show that using \methodshort~ with our family of rewriters improves performance: better reconstructions and faster convergence rates, compared with bootstrapped learning methods that do not use rewriters or use them naively.
Finally, we demonstrate that our family of rewriters can be effectively used at test time to improve the output of \methodshort~ predictions.
For 2D and 3D CSG, we outperform or match the reconstruction performance of recent domain-specific neural architectures, while producing more parsimonious programs that use significantly fewer primitives.

\end{abstract}
\section{Introduction}
\label{sec:intro}

Visual data is often highly structured:
manufactured shapes are produced by assembling parts; vector graphics images are built from layers of primitives; 
detailed textures can be created via intricate compositions of noise functions.
\emph{Visual programs}, i.e. programs that produce visual outputs when executed, are a natural approach to capturing this complexity in a structure-aware fashion.
Access to well-written visual programs supports downstream applications across visual computing domains, including editing, generative modeling, and structural analysis.
But how can we obtain a program which generates a given visual datum?
\blfootnote{Project page: \url{https://bardofcodes.github.io/coref/}}

\textit{Visual Program Inference (VPI)} methods aim to solve this problem by automatically inferring programs that represent visual inputs.
Solving this search problem is very difficult: the space of possible programs is often vast, even when constrained by a domain-specific language (DSL).
To overcome this challenge, recent works have investigated learning-based solutions, where a neural network is employed to guide the search. When a dataset of visual programs exist, such networks can be trained in a supervised fashion \cite{Wu_2021_ICCV, Fusion360Gallery, zoneGraphs, xu2022skexgen, ShapeAssembly}. Unfortunately, most domains lack such data, so recent works have investigated how to train VPI networks in an \textit{unsupervised} fashion.

\begin{figure}[t!]
	\centering
	\includegraphics[width=1.00\linewidth]{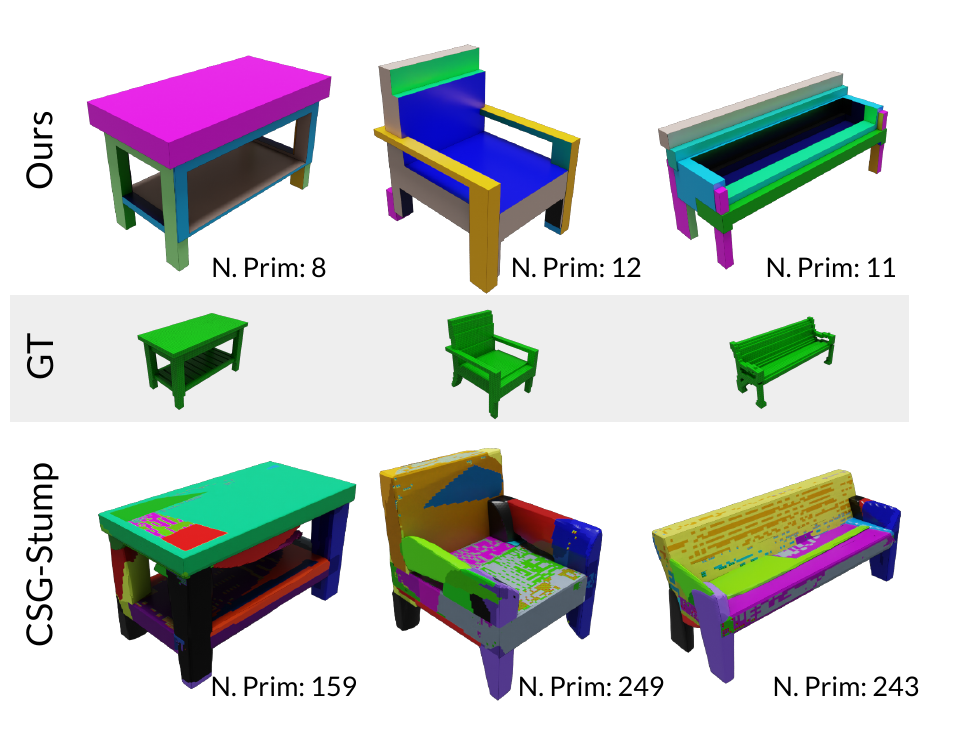}
	\caption{ 
    Our method \methodshort~ (top row) generates highly compact yet accurate programs, in contrast to CSG-Stump~\cite{CSGSTUMP_ICCV} (bottom row), which generates programs with numerous primitives. Here, we show shapes rendered with colored primitives. 
 }
\label{fig:teaser}
\end{figure}

Learning to infer visual programs without supervision is challenging: programs usually contain discrete and continuous elements, which complicates end-to-end learning.
Various solutions have been proposed to work around this issue: end-to-end learning is possible with neural architectures that act as a smooth relaxations of program executors~\cite{CSGSTUMP_ICCV}, while policy gradient reinforcement learning \cite{CSGNet} and bootstrapped learning methods \cite{jones2022PLAD} are able to treat program executors as (potentially non-differentiable) `black boxes.' 
These solutions come with downsides: designing differentiable relaxations is challenging (or even impossible) for some domains; reinforcement learning suffers from noisy gradients and slow convergence; bootstrapped learning methods are prone to getting stuck in local minima.

Moreover, a seldom acknowledged limitations of these latter methods is that they treat programs as \textit{just} sequences of tokens.
We argue that this view is suboptimal: programs are structured objects that support domain-specific reasoning to meaningfully constrain and guide the VPI search process. 
One example of such reasoning is the use of domain-specific operations that modify programs toward optimizing an objective---we call such operations \textit{rewrites}.
Rewrites have been explored in the context of VPI tasks, but primarily as a test-time optimization, e.g. finding better continuous parameters for a fixed program structure.
While such optimization is useful, our claim is that other rewrite operations are similarly useful, especially when used in tandem, and that they can be employed to benefit VPI network learning, not only as test-time optimization schemes.

In this paper, we investigate how to use code rewriting to improve visual program inference.
Unlike prior work, we focus on \textit{families} of code rewriters, each of which makes improvements to a program with some goal in mind.
We propose \methodname~(\methodshort~), a bootstrapped learning algorithm that sparsely applies rewriters and injects rewritten programs into a search-train loop at intermittent intervals.
To realize \methodshort~, we design a family of rewrites applicable to multiple visual programming DSLs: gradient-based parameter optimization (\doname), removing spurious sub-programs (\cpname), and sub-program substitutions from a cache (\cgname).  
We also propose a test-time rewriting scheme that searches for improved programs through interleaved rewrites that is well-suited to the types of programs inferred by \methodshort~-trained networks.

We evaluate \methodshort~and our family of rewriters (\doshort, \cpshort, \cgshort) on three shape program DSLs: 2D Constructive Solid Geometry (CSG), 3D CSG, and ShapeAssembly \cite{ShapeAssembly}. 
We compare VPI networks trained with \methodshort~to VPI networks trained by PLAD, a recently-proposed bootstrapped learning method \cite{jones2022PLAD}, and find that \methodshort~both increases reconstruction performance and converges significantly faster.
We further show that naively combining our rewrite families with PLAD performs much worse than \methodshort~, and in some domains even worsens performance compared with PLAD.
Finally, we demonstrate that combining \methodshort~ with our test-time rewriting scheme infers visual programs that can match (3D CSG \cite{CSGSTUMP_ICCV}) or surpass (2D CSG \cite{kania2020ucsgnet}) reconstruction performance of domain-specific neural architectures while producing significantly more parsimonious programs (see number of primitives, Figure \ref{fig:teaser}).

In summary, we make the following contributions:
\begin{packed_enumerate}
    \item \methodname~, a framework for unsupervised visual program inference that leverages a family of code rewriters.
    \item A family of code rewriters applicable to multiple DSLs that benefit VPI learning methods and can be used in a test-time rewriting scheme.
\end{packed_enumerate}

\section{Related Work}
\label{sec:rel_work}

\noindent
\textit{Visual program inference} (VPI) is a sub-problem within program synthesis.
Program synthesis is a storied field, with roots back to the inception of Artificial Intelligence, where the objective is to produce programs that meet some specification \cite{gulwani2017program}.
Under our framing, the specification is an input visual datum that the synthesized program should reconstruct when executed. 
In the rest of this section, we first overview VPI learning paradigms and then summarize prior work that looks at visual-program rewriting.

\noindent
\textbf{Learning to infer visual programs:}
\textit{End-to-end learning} methods train by propagating reconstruction loss gradients directly to a network via differentiable execution. Though such approaches can yield impressive reconstruction accuracy, they either require a soft relaxation of the program execution~\cite{kania2020ucsgnet, CSGSTUMP_ICCV, Yu_2022_CVPR_caprinet, dualCSG, ren2022extrude, carlier2020deepsvg} which is infeasible for many languages,
or require training domain-specialized neural executors~\cite{tian2019learning, hu2022diff}, which can introduce approximation errors. 
In \methodshort~, we instead leverage a `partially' differentiable execution of visual programs, bypassing the need of program relaxation or neural executors. 

\textit{Reinforcement learning} has also been used by prior VPI approaches~\cite{CSGNet, ellis2019write, abstractionTulsiani17}. Usually, the inference network is treated as an `agent' maximizing a reward signal tied to its reconstruction accuracy. 
The high variance of policy gradient methods has limited the application of such techniques to toy datasets, especially for 3D data. 
Similar in spirit to \methodshort~, other programmatic RL methods for non-visual domains have explored blends of program optimization and learning ~\cite{propel_RL, PiRL}.
Also related are non-programmatic RL methods that explore episode modification, through episode relabeling or neurally-guided search ~\cite{NIPS2017_HER, iHER, fang2018dher, NIPS2018_MAPO, counterfactual_her}.
Like \methodshort~, these approaches aim to improve learning targets through local search, but they do so for vastly different domains (often much simpler than complex 3D shape-programs),  employ simplistic rewriting techniques, and don't target bootstrapped learning frameworks.

\begin{figure*}[t!]
    \centering
   \includegraphics[width=1.0\linewidth]{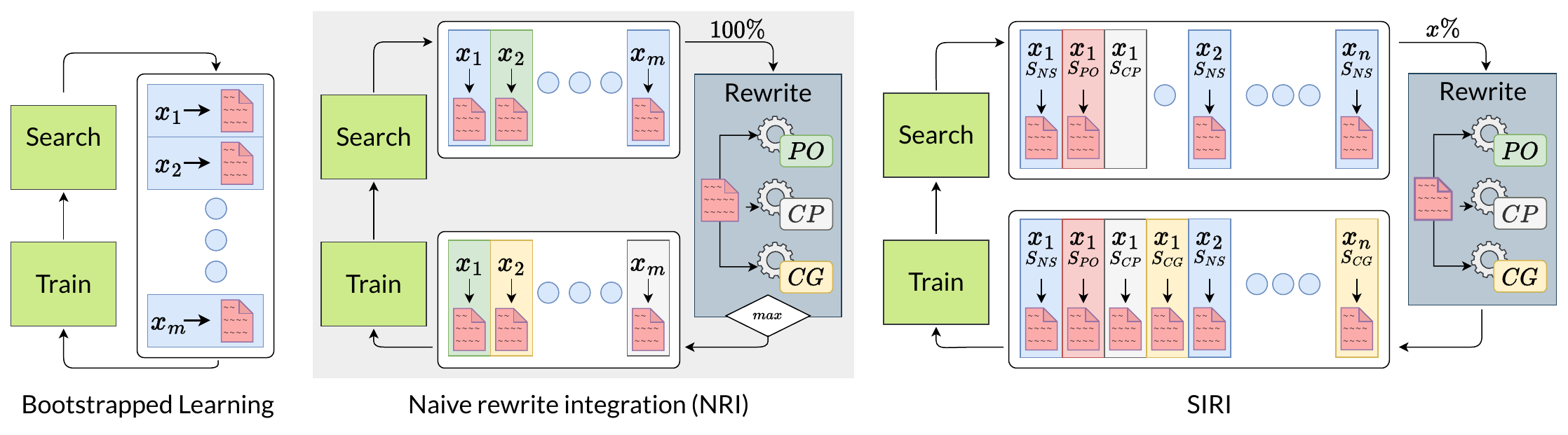}
    \caption{
 Bootstrapped learning methods~\cite{jones2022PLAD} store a record of the objective maximizing program for each shape in a Best Program (BP) data-structure. 
 Naive rewrite integration applies a family of rewriters to each entry in BP, and overwrites each entries when a rewrite is successful. Our method \methodshort~, instead applies the rewriters to a subset of BP entries and overwrites entries only when their sources $S_i$ match. 
    }
    \label{fig:bootstrapped_learning} 
\end{figure*}

\textit{Bootstrapped learning} is an attractive alternative that can avoid the pitfalls of RL and end-to-end learning ~\cite{jones2022PLAD, liang2017_ILM}. 
Such approaches alternate between \search phases, that discover `good' programs, and \train phases, that use discovered programs to train a network.
While these approaches have demonstrated improvements over RL, they are still limited by treating each program as a sequence of tokens, rather than a structured object. 
Our work bridges this gap by offering an effective way of integrating a family of code rewriters into a recent bootstrapped learning paradigm \cite{jones2022PLAD}.

\noindent
\textbf{Rewriting visual programs:} 
\textit{Gradient-based optimization} is a common approach for optimizing visual programs. 
For neural architectures that serve as a relaxation of the executor, this can be achieved via test-time fine-tuning.
While such approaches achieve impressive reconstruction performance, they are expensive to run and often produce messy program structure ~\cite{Yu_2022_CVPR_caprinet, dualCSG}. 
This fine-tuning can take prohibitively long to converge, requiring anywhere from $3$ minutes~\cite{Yu_2022_CVPR_caprinet} to even 30 minutes~\cite{CSGSTUMP_ICCV} \textit{per sample}.

Typically, visual program executors can be made piecewise differentiable with respect to parameters of an input program (up to control flow decisions), which supports test-time gradient-based optimization ~\cite{CSGNet, sharma2020parsenet, materials_match}.
Yet, as this rewriting scheme does not change program structure, reliance on \textit{only} gradient-based optimization is vulnerable to getting stuck in local minima. 
We employ a \doname rewriter (Sec. \ref{sec:met_DO}) as one member of a rewrite family, where other rewriters can make structural program changes, to avoid getting stuck in these local minima.
Critically, our implementation is highly-efficient, and we can apply each member of our rewrite family multiple times in under 20 seconds during test-time optimization.

\textit{Gradient-free optimization} techniques have been investigated that leverage domain heuristics to develop rewriting strategies that modify the control flow decisions of visual programs. 
For 3D CAD languages, these operations have been explored for non-learning based reverse-engineering methods ~\cite{InverseCSG, 2018-icfp-reincarnate}.
Recently, E-graphs \cite{egg} have been employed to efficiently search for rewritten programs that optimize criteria such as program length \cite{szalinski} or fabrication cost \cite{zhao2021co,wu2019carpentry}. 
While these methods do not consider \textit{learning} from rewritten programs, it would be possible to integrate these types of techniques into our family of rewriters. 

\textit{Abstraction discovery} is a special form of rewriting, where common subcomputations shared across many programs are factored out into subroutines (i.e. abstractions), and programs are rewritten to use these subroutines.
When a dataset of programs is given as input, this step can be decoupled from visual program inference \cite{jones2021shapeMOD, yang2022discovering, babble, topdownlib}.
Alternatively, some methods have investigated how abstraction discovery (AD) phases can improve visual program inference performance~\cite{DreamCoder, ellis2018library}. In an iterative procedure, an abstraction phase greedily rewrites a dataset of programs with abstractions, then a recognition model learns on rewritten programs to discover higher-order abstractions and solve more complex inference problems.
Such methods are not yet able to scale to the complex 3D shape domains we study in this work, as they employ simplistic recognition models
and rely on expensive enumerative search.
Furthermore, we find that naively integrating rewriter outputs, as done in past AD approaches, can in fact be detrimental to the bootstrapping process. 
We instead propose \methodshort~, a non-deleterious procedure for integrating rewriters under bootstrapped learning paradigms, described in Sec. \ref{sec:met_bs_learn}, which may prove similarly beneficial for AD approaches.

\section{Method}
\label{sec:method_split}

In this section, we explore how families of code rewriters can be employed to improve visual program inference.
In section \ref{sec:met_task_spec}, we formalize our task specification, objective function, and rewriter assumptions.
We then present \methodname~(\methodshort~), an unsupervised learning paradigm for visual program inference (VPI) in Section \ref{sec:met_bs_learn}. 
\methodshort~ employs a family of rewriters to improve reconstruction performance for VPI tasks while maintaining a parsimonious program representation.
Finally, we describe how these operations can also be used in a test-time rewriting scheme, that is especially well-suited to the outputs of \methodshort~(Sec. \ref{sec:met_ttr}). 
We describe the family of rewriters we employ for shape-program domains in Section \ref{sec:method_rewrite}.

\subsection{Task Specification}
\label{sec:met_task_spec}

We define the visual program inference task as follows:
given a target distribution $S$ of visual inputs (e.g. shapes), we want to learn a model $p_\theta(z|x)$, where $x \in S $, which infers a program $z$ whose execution $E(z)$ reconstructs the input shape $x$.
Following Occam's razor, we seek programs that are parsimonious. More formally, we seek a $p_{\theta^*}$ that maximizes our objective function \Obj:
\begin{align}
\label{eq_overall}
    \theta^* &= \argmax_{\theta} \bigE_{x \in S} \biggl[ \sum_{z} \mathcal{O}(x, z) p_\theta(z|x)  \biggr], \\
\label{eq_objective}
    \mathcal{O}(x, z) &= \mathcal{R}(x, E(z)) - \alpha |z|,
\end{align}
where $\mathcal{R}$ measures the reconstruction accuracy between $x$ and $E(z)$, $|z|$ measures the length of the program, and $\alpha$ modulates the strength of these two terms. 
Our method takes in a family of rewriters, \RWS, to help with this task.
Though each rewriter may leverage different domain properties, they are all tied to the same objective \Obj. 
Formally, they aim to perform rewrites \RW$(x,z)\rightarrow z_i^R$ s.t. $z_i^R \sim \argmax_{z\in \Omega^i(z)} \mathcal{O}(x, z) $, where $\Omega^i(z)$ represents the set of all programs rewriter $i$ can create from $z$.

\subsection{Sparse Intermittent Rewrite Injection}
\label{sec:met_bs_learn}

One way to use rewriters to improve VPI models is to incorporate them into bootstrapped learning schemes. 
One such recent bootstrapped learning paradigm is proposed by PLAD \cite{jones2022PLAD}. 
PLAD alternates between search and train phases, where a best-program (BP) data-structure modulates how these phases communicate (Figure~\ref{fig:bootstrapped_learning}, left). 
In the search step, a VPI network $p(z|x)$ is fed visual inputs from $S$ and aims to predict programs that optimize \Obj (e.g through beam-search). 
The results of this neurally-guided search (NS) are used to populate the fields of BP, where each key corresponds to a unique shape $x$, and each value corresponds to the best program (in terms of \Obj, for $x$) seen so far.
In the train step, the entries of BP are used to construct paired training sets $(X, Z)$ which are then used to optimize $p_\theta(z|x)$ with maximum likelihood loss.

How can rewrites be integrated into such a framework? We demonstrate a naive approach, termed \pplusr (\pplusrshort), in Figure~\ref{fig:bootstrapped_learning}, middle. 
In this naive version, a \rewrite step takes place between each search and train step, that modifies the entries of BP. 
Specifically, for every $(x,z)$ entry in BP, each rewriter is applied to $z$, and if $z^R$ improves the \Obj, then $(x, z^R)$ is entered into BP, overwriting the $(x, z)$ entry. 
Empirically, we find that this approach can in fact be worse than PLAD.
While the rewriters perform a local search to optimize \Obj for the given $(x, z)$ pair, there is no guarantee that a discovered $z^R$ will make a better training target for $p_\theta(z|x)$ as well.
\pplusrshort has a deleterious effect on the entries of BP, as only one program value is maintained for each shape key, so indiscriminate applications of rewriters can get easily stuck in local minima with respect to $p(z|x)$. 
Prior works which incorporate program rewriting strategies~\cite{ellis2018library, DreamCoder} update training programs in this way.

\methodname~ (\methodshort~) also employs a \rewrite step, but avoids this issue with a more judicious application of rewriters from \RWS (Figure~\ref{fig:bootstrapped_learning}, right). Instead of applying rewriters on each shape, rewrites are only applied to a subset of BP entries. \methodshort~ also adds a source field to each BP key that indicates what produced a given program: either neurally-guided search (NS) or some \RW~$\in$ \RWS~.
Critically, this allows \methodshort~to add programs into BP in a way that can only `forget' (shape, program) pairs from the same source: NS programs replace NS programs, and each \RW~can only bump out \RW~sourced entries.

During the search-step, each $(x, z)$ pair produced by the neurally-guided search populates the $(x, S_{NS}, z)$ entry of BP.  
Then for each \RW~$\in$ \RWS~, \methodshort~ samples a percentage of BP entries, $(x, src, z)$, and adds $(x, \RW, z^R)$ if the rewriter successfully improves the objective. Note that rewrites can sample inputs from any source, not just $S_{NS}$. Not only does this scheme ensure that rewriters only overwrite their own previous predictions, but it is also more efficient versus \pplusrshort: some rewriters have high computational costs, and excessively applying them can slow down training.
Instead, we find that training $p(z|x)$ on BP entries with sparsely rewritten programs, both converges faster and reaches better final end-states, as useful rewriting strategies get amortized by the network weights. 

As only a subset of entries are updated in each \rewrite phase, some rewrite entries in BP (potentially from previous \rewrite phases) may store programs which are worse than the program inferred by $p_\theta(z|x)$ during \search.
Therefore, before each train step, \methodshort~purges all stale $(x, \RW, z^*)$ entries from BP whenever $\mathcal{O}(x, z^*) < \mathcal{O}(x, z)$, where $z$ is the program inferred for $x$ during the \search phase.

\begin{figure*}[t!]
	\centering
 
	\includegraphics[width=1.0\linewidth]{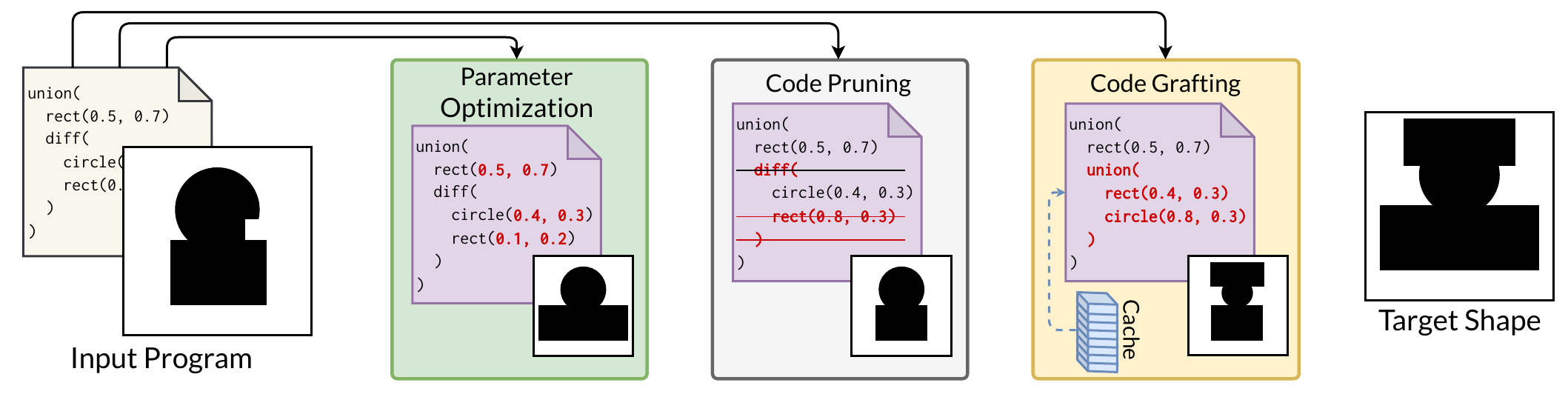}
	\caption{
	\methodshort~uses \textit{rewrites} to improve VPI networks. 
	We depict three rewrites in action that: 
	optimize continuous parameter values (\doname), 
	remove extraneous code (\cpname),
	and substitute sub-expressions from a cache (\cgname).
 }
	\label{fig-rewrite-overview}
\end{figure*}

\subsection{Test-time Rewriting}
\label{sec:met_ttr}

Our family of rewriters, \RWS can also be employed at inference time to find better programs for a particular visual target.  
With a slight abuse of notation, given an $(x, z)$ pair, our test-time rewrite (TTR) approach aims to find a sequence $seq^*$ of at-most $k$ rewrite operations from \RWS, such that 
 $seq^* \sim \argmax_{seq \in \RWS^k} \mathcal{O}(x^*, seq(x, z))$, where $z^{ttr} = seq^*(x, z)$ would be the result of applying each rewrite in $seq^*$ to $z$ iteratively.

We realize this formulation with a greedy search. 
We initialize $z^{ttr}$ as $z$, and then for $k$ steps, we iterate through each \RW~$\in$ \RWS, compare \Obj$(x,z^{ttr})$ to \Obj$(x,\RW(z^{ttr}))$, and replace $z^{ttr}$ with $\RW(z^{ttr})$ if \Obj improves.
To avoid redundant work, if a \RW has already tried to improve a particular $z$, and failed to do so, we instead pass in the next best observed $z$ (in terms of \Obj) that the \RW has not operated over.
The final $z^{ttr}$ is then returned as the output.

This procedure can in theory be applied to programs produced by any source (e.g. networks that acts as a differentiable language executor~\cite{CSGSTUMP_ICCV}).
However, we find that there are unique benefits to applying this procedure to the predictions made by a $p_\theta(z|x)$ network trained with \methodshort~. Some rewrites build a cache of partial results during bootstrapped learning that can be quickly and effectively applied at inference time (\cgshort, Section~\ref{sec:met_CG}). Furthermore, some rewrites are too expensive to run on very complex programs, consuming massive amounts of memory, but are well-suited to the parsimonious programs produced by \methodshort~. For instance, one rewriter (\doshort, Section~\ref{sec:met_DO}) was not able to operate on the highly complex programs output by CSG-Stump~\cite{CSGSTUMP_ICCV} (cf. supplemental material).

\section{Rewriting Visual Programs}
\label{sec:method_rewrite}

As our method relies on an input family of rewriters, \RWS, we identify three rewrite operators that generalize across multiple shape-program domains.
Figure \ref{fig-rewrite-overview} depicts these rewriters in action.
In the rest of this section, we provide a high-level description and motivation for the different rewriters we use during \methodshort~ and test-time rewriting: \doname (Section \ref{sec:met_DO}), \cpname (Section \ref{sec:met_CP}), and \cgname (Section \ref{sec:met_CG}). 
We provide the implementation details in the supplemental material.

\subsection{Parameter Optimization}
\label{sec:met_DO}

Visual languages often contain continuous (and differentiable) parameters such as the scale and position of primitives, and discrete parameters such as control flow (e.g. how to combine primitives). 
While keeping the discrete parameters fixed, \doname(\doshort) rewriter aims to improve the continuous parameters of a given program using gradient-based optimization.
Given a program $z_\phi$ with continuous parameters $\phi$ inferred for a shape $x \in S$, \doshort adjusts $\phi$ to maximizes the reconstruction accuracy $\mathcal{R}$ between $x$ and the program execution $E(z_\phi)$: $\phi^* \sim \argmax_\phi \mathcal{R}(x, E(z_\phi))$.

To propagate gradients back from a reconstruction metric to the continuous parameters of $z_\phi$, \doshort requires that the program executor $E$ is partially differentiable: one or more continuous parameters should have well-defined derivatives for a given program structure (though the program execution may itself be only piecewise continuous). 
We highlight that \doshort is useful \textit{even} under such constraints, precisely because \doshort is not the only rewriter we consider: other rewriters in \RWS can influence structural changes, along with \methodshort~'s \search phase.
\doshort's requirements on $E$ differ significantly from the differentiable executors employed by neural relaxation architectures. 
These works~\cite{kania2020ucsgnet, CSGSTUMP_ICCV} often attempt to differentiate through both discrete and continuous decisions, which leads to noisy gradients and poor program quality. 

Although the design of executors is domain-dependant, we outline our procedure for converting the outputs of $E$ into an equivalent signed-distance field representation compatible with our reconstruction metric $\mathcal{R}$. 
For each language we consider, we map outputs from $E$ into a tree-like representation, where each leaf represents a primitive (spheres, etc.) and each intermediate node represents a transformation (position, etc.) or a combinator (union, etc.). 
For CSG-like languages, we can directly map from program parameters $\phi$ to this representation. 
It is also possible to convert the output of more complex program executors that produce collections of primitives into this format \cite{ShapeAssembly}. 
Then, we perform boolean combinations of the parameterized primitives, and apply the transformation operators, to obtain the program's implicit equivalent. 
With the program's implicit equivalent, we now uniformly sample points $t \in \mathbb{R}^n$, and convert the signed distance at the points into soft-occupancy to yield a differentiable execution of the program.
This framing should be extensible to other visual-programming domains of interest: where either this mapping may be explicitly extracted from the input program (SVG) or parsed from primitives created by a more complex executor \cite{pearl2022geocode}.

\subsection{Code Pruning}
\label{sec:met_CP}

One drawback of bootstrapping techniques is their tendency to reinforce spurious patterns~\cite{liang2017_ILM}. 
The \cpname (\cpshort) rewriter mitigates this problem by identifying and removing program fragments that negatively contribute to our objective $\mathcal{O}$.
Given a shape $x$ and input $z$, \cpshort rewrites $z$ s.t. $z^R \sim \argmax_{\Omega^{CP}} \mathcal{O}(x, z)$, where $\Omega^{CP}(z) = \{z* | z* \subseteq z \}$ represents the set of all valid sub-programs of $z$.
A naive \cpshort rewriter that considers every valid sub-expression would be prohibitively slow.
We instead implement a greedy version of \cpshort designed for declarative, functional languages.  
We describe our general approach below, and provide further details in the supplemental material. 

Our implementation of \cpshort employs two greedy searches, a top-down pass and a bottom-up pass, to approximate $z^R$. At each pass, we identify and prune nodes which decrease the overall objective score \Obj. The top-down traversal relabels the highest objective scoring node as the root, pruning all but the tree starting at that node. The bottom-up traversal then checked each node's contribution to the final execution and prunes branches with negligible contribution. 

\begin{figure}[t!]
	\centering
	\includegraphics[width=1.0\linewidth]{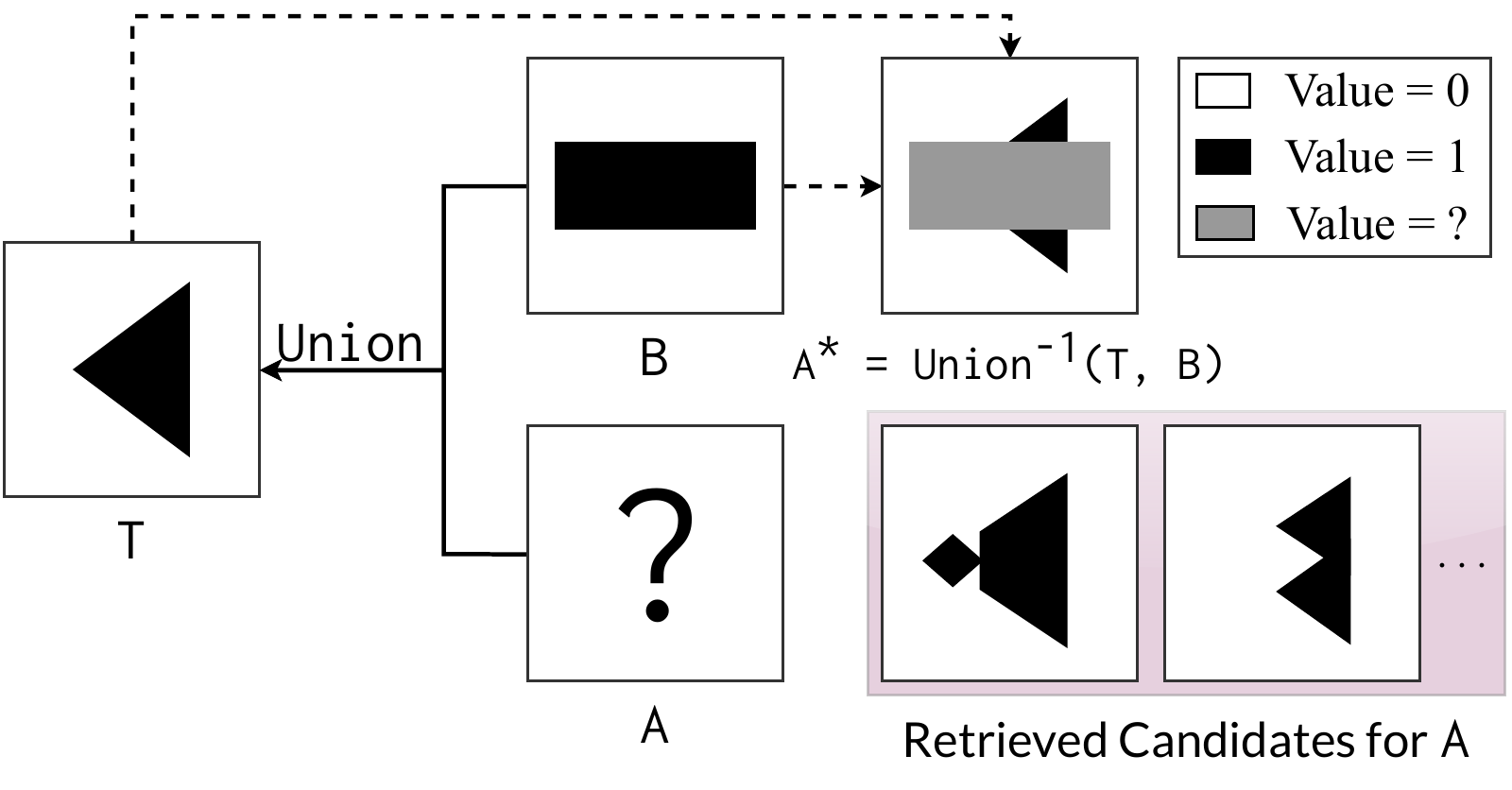}
	\caption{
	Given a target $T$, we derive the \textit{desired execution}, $A^*$, for sub-expression $A$, through operator inversion. 
	Our \cgshort rewriter leverages \textit{desired-executions} to search for replacement candidates.
	}
	\label{fig-inversion}
\end{figure}

\subsection{Code Grafting}
\label{sec:met_CG}

A key feature of symbolic representations is the ability to perform part-based reasoning, e.g. compose parts from different instances. The \cgname (\cgshort) rewriter exploits this feature to improve programs. Specifically, the \cgshort rewriter replaces sub-expressions of a particular program with a better sub-expression (in terms of \Obj). 
The primary challenge is specifying \emph{what} to replace a sub-expression with, as the space of potential replacements is enormous.
To make the search tractable, \cgshort builds a program cache populated by sub-expressions discovered while running \methodshort~, and searches for replacement candidates within the cache. 

When the cache is large, searching for good replacement expressions can become expensive.
Our \cgshort implementation makes this search tractable by focusing on \textit{execution equivalence}, i.e. equivalence is measured by comparing their executions (stored as an $n$-dimensional occupancy grid). 
But finding cache entries which would improve reconstruction performance requires some notion of the \textit{desired execution}, i.e. what sub-expression execution would make the program better match the target shape $x$?

We develop a procedure for calculating such desired execution through masked function inversion; an example is depicted in Figure \ref{fig-inversion}. 
We provide the high-level insight by walking through this figure; further details are in the supplemental material.
In this example, $T$ denotes the target shape (the desired final execution), which is a union of two subexpressions.
Suppose we wish to replace subexpression $A$.
Given the current state of its sibling subexpressions ($B$, in this case), we can invert the Union to produce the \textit{desired-execution} $A^*$. 
$A^*$ can be broken into sub-regions: (black and white) areas where the optimal execution behavior is known and (gray) areas where the optimal execution behavior is unknown (for example, due to the non-invertibility of some operators). 
\cgshort uses such ternary \textit{desired-executions} to search for cache entries that are likely to improve reconstruction accuracy and substitutes the most suitable candidate.

\section{Results}
\label{sec:experiment}
We evaluate the efficacy of our rewriters over a collection of VPI domains for two tasks: improving bootstrapped learning (Section~\ref{sec:exp_blend}), and improving VPI with test-time rewrites (Section~\ref{sec:exp_ttr}). First, we provide the details concerning our experiments in Section~\ref{sec:exp_des}.

\begin{table*}[t!]
\small
\begin{center}
\begin{tabular}{lcccccc}
    \toprule
    & \multicolumn{3}{c}{Chamfer Distance ($\downarrow$)} & \multicolumn{3}{c}{IoU ($\uparrow$)} \\
    \midrule
    & 2D CSG & 3D CSG & {\small ShapeAssembly} & 2D CSG & 3D CSG & {\small ShapeAssembly}\\
    \midrule
    PLAD~\cite{jones2022PLAD}   & $0.24$ & $1.75$& $1.98$ & $90.8$ &$74.65$& $63.1$\\
    \pplusrshort  & $0.36$ & $1.25$ & $1.53$ & $88.4$& $74.43$ & $66.89$\\
    \midrule
    \methodshort~& $\mathbf{0.22}$ & $\mathbf{1.10}$ & $\mathbf{1.44}$ & $\mathbf{91.7}$& $\mathbf{76.77}$ & $\mathbf{67.8}$\\
    \bottomrule
\end{tabular}
\end{center}

\caption{
We report the Test-set performance across 3 VPI domains. Naively integrating the rewriters into PLAD (\pplusrshort) can detoriorate the model's performance (IoU on 2D \& 3D CSG). In contrast, \methodshort~ consistently improves over PLAD on all the three VPI domains.
}
\label{table:ablation}
\end{table*}

\subsection{Experimental Design}
\label{sec:exp_des}

\noindent
\textbf{Domain-Specific Languages:} 
We consider three VPI domains: 2D Constructive Solid Geometry (CSG), 3D CSG, and ShapeAssembly \cite{ShapeAssembly}.
Shapes are formed in CSG by declaring primitives such as cylinders, applying transformations, and composition via boolean operations.
ShapeAssembly produces hierarchies of cuboid part proxies (which can themselves contain sub-programs) assembled through attachment operations. Please see the supplemental for the complete DSL grammars. 

To ease learning, past approaches have used simplified versions of these languages, e.g. restricting CSG to contain only primitive-level transformations or removing hierarchical sub-programs from ShapeAssembly \cite{kania2020ucsgnet, jones2022PLAD, CSGNet}. For fair comparison, we match our DSLs to prior work. 

\noindent
\textbf{Shape Datasets:} We evaluate 2D CSG on the CAD dataset introduced in CSGNet~\cite{CSGNet}. It contains front and side views of chairs, desks and lamps from the Trimble 3D warehouse. This dataset is divided into 10K training, 3K validation and 3K testing shapes. We evaluate 3D CSG and ShapeAssembly on the 3D CAD dataset released in~\cite{jones2022PLAD} containing shapes from chair, table, couch, and bench categories of ShapeNet dataset~\cite{chang2015shapenet} in a voxel grid format. This dataset is split into 10k training, 1k validation and 1k testing shapes.

\begin{figure}[t!]
    \centering
   \includegraphics[width=1.0\linewidth]{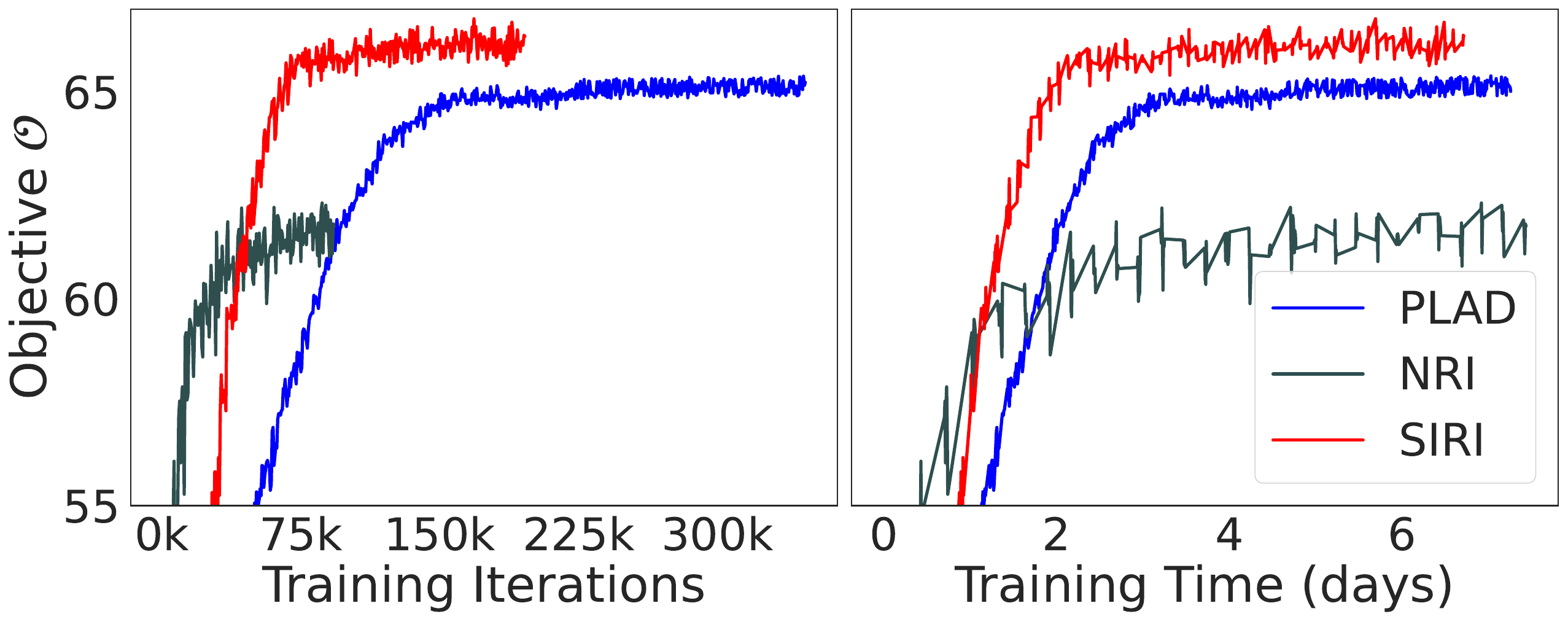}
    \caption{
    We plot the objective $\mathcal{O}$ (Y-axis) as a function of training time (X-axis), measured by iterations (left) and wall-clock time (including time taken for rewriting) (right) on 3D CSG domain.
    By amortizing the rewrite-cost, and keeping the training data diverse, \methodshort~ achieves higher performance and does so faster.
}
    \label{fig:training_speed} 
\end{figure}

\noindent
\textbf{Model Architecture:} Our model $p(z|x)$ synthesizes programs as a sequence of tokens, where each token specifies a command type or its parameters. Numeric parameters are normalized and discretized into 33 bins.
Each $p(z|x)$ uses a domain-specific feature extractor (e.g. a 2D or 3D CNN), and a decoder-only transformer module. 
For 2D languages, the feature extractor takes a $64^2$ image as input; for 3D languages, it takes a $32^3$ voxel grid. We use the same transformer architecture for all experiments, varying only the last layer output size to model the different number of commands in each language. 

\noindent
\textbf{Metrics:} 
We measure reconstruction accuracy with two metrics: Intersection Over Union (IoU) and point cloud Chamfer-Distance (CD). We use $64^2$ and $32^3$ resolution occupancy grid for calculating 2D and 3D IoU respectively. We follow the same methodology as CSGNet~\cite{CSGNet} for measuring 2D CD; 3D CD is measured between $2048$ points sampled on the ground-truth ShapeNet meshes and the meshes produced by executing the inferred programs. 

\noindent
\textbf{Training details:} Following~\cite{CSGNet, jones2022PLAD, ellis2019write}, we pretrain our models on a large corpus of synthetically generated programs until it converges. We generate the synthetic programs via the sampling procedure proposed in PLAD~\cite{jones2022PLAD}.
After pretraining, the model is finetuned on the target distribution $S$, following the procedure outlined in Section~\ref{sec:met_bs_learn}.
During each \rewrite phase, we apply \doshort, \cpshort and \cgshort to $50\%$, $50\%$, $15\%$ of programs respectively. \cgshort is applied to only $15\%$ of data due to its higher computational cost.
For our training objective $\mathcal{O}$ (c.f. equation~\ref{eq_objective}), we fix $\alpha$ to $0.015$. For 3D we set $\mathcal{R}$ to IoU, and for 2D we set $\mathcal{R}$ to CD. 
For test-time rewriting (TTR), we perform interleaved application of \textit{each} rewriter thrice, unless specified otherwise (i.e. for Table~\ref{table:ttr_sequential}).

\subsection{Training with SIRI}
\label{sec:exp_blend}

We first evaluate the benefit of intermittent rewriting for bootstrapped learning. We compare our method \methodshort~ against 2 baselines, namely PLAD~\cite{jones2022PLAD}, and \pplusr~ (\pplusrshort) which naively integrates the rewriters into PLAD (cf. Section~\ref{sec:exp_blend}). 
Note that prior work on integrating code rewriting~\cite{ellis2018library, DreamCoder} follow this strategy.

\begin{figure}[t!]
    \centering
   \includegraphics[width=1.0\linewidth]{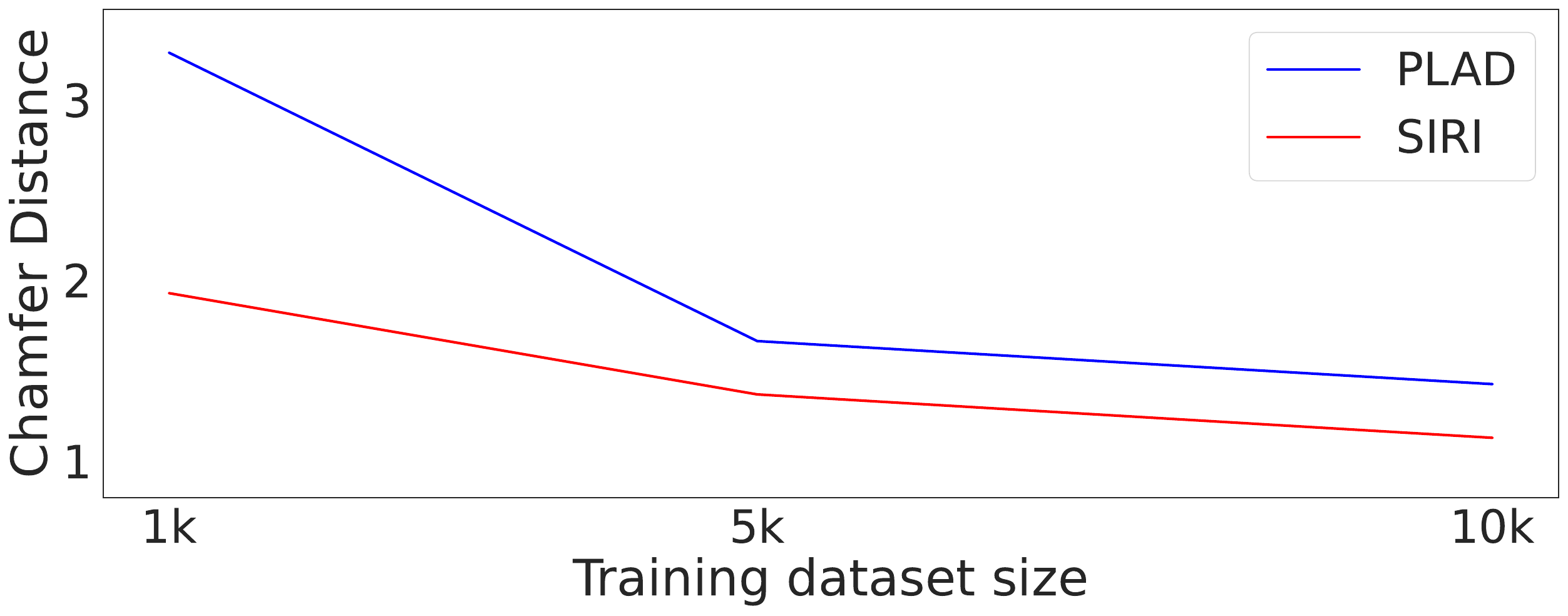}
    \caption{
    Due to data-independent rewriters, \methodshort~ remains effective at training the network even with a fraction of the data. In comparison, since PLAD's \search phase is tied to the inference network's performance, data scarcity deteriorates its performance. 
    }
    \label{fig:data_scacity} 
\end{figure}

As shown in Table~\ref{table:ablation}, \methodshort~ outperforms both the baselines on all domains. While PLAD performs well on simple domains such as 2D CSG, it is less effective on harder domains such as ShapeAssembly. Naively integrating the rewriters (\pplusrshort) can in fact even be detrimental to bootstrapped learning, as can be seen for 2D \& 3D CSG (w.r.t. IoU).
Excessive use of the rewriters and the lack of training data diversity leads the model trained with \pplusrshort to overfit to a local minima, i.e. the \search and \rewrite phases become ineffective at generating useful better programs to learn from.
\methodshort~ resolves this issue by its frugal usage of the rewriters and by training $p_\theta$ on a diverse set of programs obtained from both the \search and \rewrite phases.
We see a similar trend on other visual languages, which we discuss in the supplemental material. 
Finally, we note that with $0.22$ CD on 2D CSG domain, \methodshort~ out-performs UCSG-Net~\cite{kania2020ucsgnet} ($0.32$ CD), the previous state-of-the art method for 2D CSG.

We also evaluate how \methodshort~ impacts the training convergence in Figure~\ref{fig:training_speed}, plotting the Objective $\mathcal{O}$ against iterations (left) and wall-clock time (right).
Wall-clock time includes the time required to execute the rewriters.
Though \pplusrshort starts with a high performance, it eventually converges to a lower value of $\mathcal{O}$ despite expending a lot of time for rewriting. In contrast, \methodshort~ is able to achieve a higher performance while also amortizing the cost of rewriting all the programs as the model generalizes the useful patterns present in the rewritten training programs. 

\noindent
\textbf{Data Scarcity}: Often, large datasets of example shapes are not easily available. Thus, we probe the efficacy of \methodshort~ and PLAD under data scarcity. 
We present our experiment in Figure~\ref{fig:data_scacity}, plotting the validation set CD for PLAD and \methodshort~.
With 100\% data, \methodshort~ surpasses PLAD by $0.29$ CD, where as at 10\% data, \methodshort~ outperforms PLAD by a margin of $1.33$ CD.
Since PLAD relies solely on neurally-guided search to discover good programs, which is dependant on the dataset size, reduction in dataset size hurts their performance. In contrast, as \methodshort~ employs rewriters such as \doshort and \cpshort which are invariant to the training dataset size, it outperforms PLAD in low-data regimes.  

\subsection{Test-Time Rewriting}
\label{sec:exp_ttr}

We now show how combining \methodshort~ with additional rewriting at test-time allows performance that matches state-of-the-art methods specialized for 3D CSG reconstruction, while producing much more parsimonious programs.
Specifically, we now compare \methodshort~ against CSG-Stump~\cite{CSGSTUMP_ICCV}, a state-of-the-art method with a neural architecture designed specifically for CSG reconstruction.

\begin{table}[t]
\small
\begin{center}
\begin{tabular}{lccc}
\toprule
& {\small CD ($\downarrow$)} & {\small N. Prim. ($\downarrow$)} & {\small N. ops. ($\downarrow$)}\\
\midrule
CSG-Stump 32 & $1.90$ & $19.03$ & $5.95$ \\
CSG-Stump 256 & $1.22$ & $154.47$ & $52.85$\\
CSG-Stump 256 (cs) & $0.78$ & $191.38$ & $72.52$\\
\midrule
 \methodshort~ & $1.101$ & $3.90$ & $2.90$ \\
 \methodshort~ + TTR & $0.83$ & $8.47$ & $7.42$ \\
\bottomrule
\end{tabular}
\end{center}

\caption{
\methodshort~ outperforms CSG-Stump 256 with a fraction fewer primitives. Applying test-time-rewrites to \methodshort~ makes its performance comparable to that of class-specific (cs) CSG-Stump while remaining relatively parsimonious. 
}
\label{table-3dcsg_comparison}
\end{table}

We train CSG-StumpNet with the author-released code on our dataset. 
We compare against two versions: CSG-Stump 32 with 32 intersection and union nodes and CSG-Stump 256 with 256 intersection and union nodes. Since the authors originally trained their model independently for each class, we also compare against a class-specific (cs) version of CSG-Stump 256, where we use pretrained class-specific models released by the authors.

We show our results in Table~\ref{table-3dcsg_comparison}. We see that \methodshort~ achieves $0.1$ CD lower than CSG-Stump 256, despite being domain agnostic. More importantly, it achieves this with a fraction of primitives and operations. 
Secondly, when accompanied by test-time rewrites, \methodshort~ achieves similar CD to CSG-Stump 256 (cs), which has models trained \textit{individually} for each class. \methodshort~ achieves this while having an order of magnitude fewer primitives and operators (yielding more interpretable and editable programs).
In Figure~\ref{fig:teaser}, we visualize the difference in inferred program size by rendering their executed shapes with colored primitives. 

As the programs inferred by \methodshort~ are parsimonious, applying TTR to the inferred programs takes only $14.6$ seconds per shape on an average. In contrast, the over parameterized programs inferred by CSG-Stump are not amenable to fast test-time-optimization (cf. supplemental).
Instead, prior-works~\cite{Yu_2022_CVPR_caprinet, dualCSG} fine-tune the network itself for each test-shape, which, for CSG-Stump, requires $\sim$ 30 minutes per shape~\cite{dualCSG}. Moreover, while network fine-tuning can increase reconstruction accuracy, the inferred programs remain incomprehensibly large.

\begin{figure}[t!]
	\centering
 
	\includegraphics[width=1.0\linewidth]{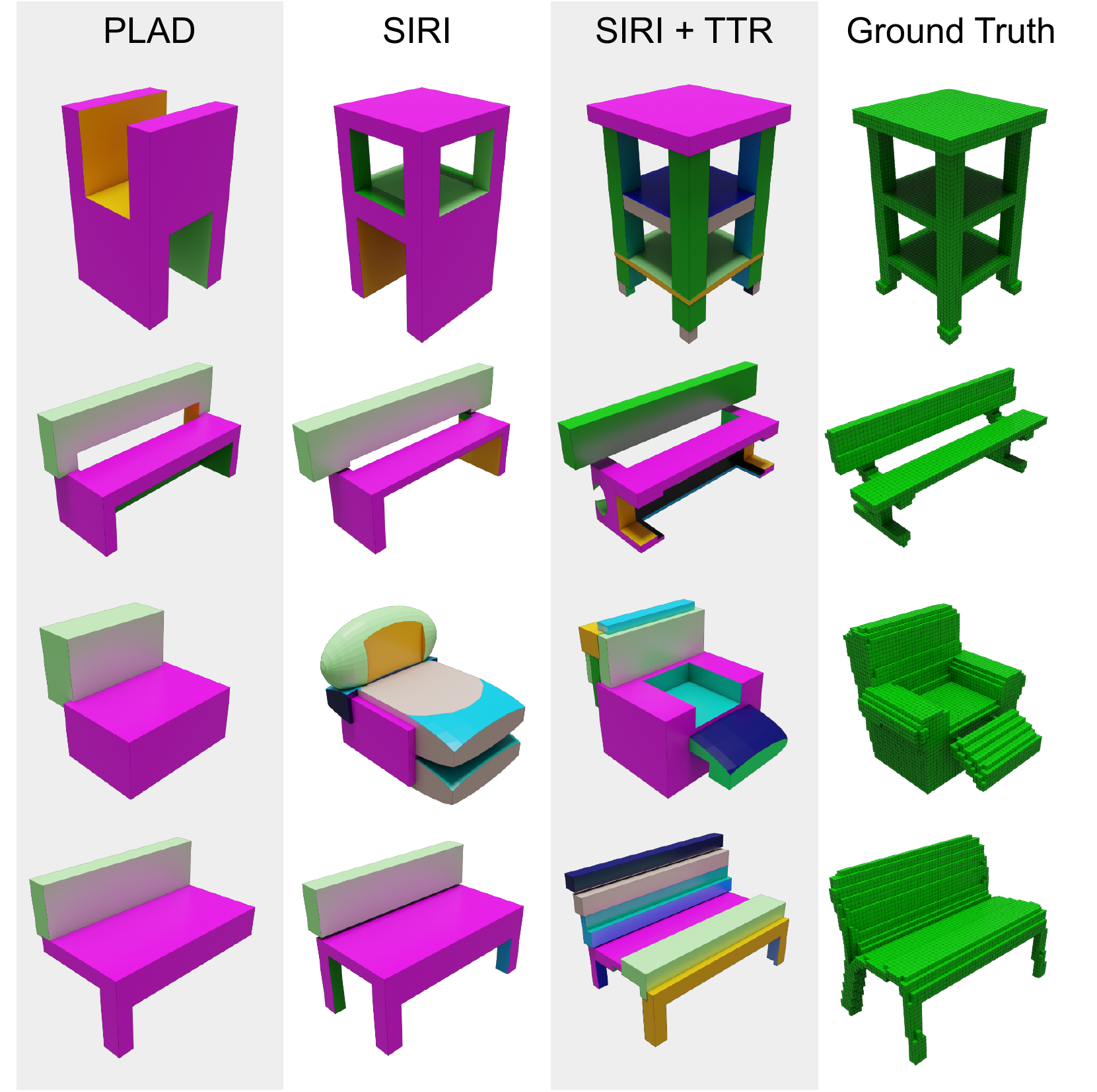}
	\caption{
    We present qualitative examples of our method and PLAD~\cite{jones2022PLAD}. \methodshort~ outperforms PLAD and test-time rewriting improves it further.
 }
	\label{fig:qualitative}
\end{figure}

Test-time rewrites are beneficial for PLAD inferred programs as well. However, we find it best to both (i) train with rewriters and (ii) use them at inference time.
We compare test-time rewrites on models trained with \ablation and \methodshort~ and report the results in Table~\ref{table:ttr_on_plad}. 
For both 3D CSG and ShapeAssembly, using test-time rewrites on \methodshort~ is more beneficial than using them on PLAD. Apart from yielding better initialization for the inferred programs, training with \methodshort~ also equips the \cgshort rewriter with a program cache filled with useful sub-expressions; this cache bolsters \cgshort's efficacy at test-time rewriting. 
Note that \methodshort~ + TTR programs have only a marginal increase in length over PLAD + TTR.

As described in Section~\ref{sec:met_ttr}, we interleave the application of our three rewriters for test-time rewriting, allowing changes to both the structure and continuous parameters of the program. In Table~\ref{table:ttr_sequential}, we evaluate the impact of each individual rewriter.
Both \doshort and \cgshort improve upon the program inferred by beam search while taking only a few seconds. Combining them further improves performance while keeping the programs relatively parsimonious.

\begin{table}[t!]
\small
\begin{center}
\begin{tabular}{lcccc}
    \toprule
    &  \multicolumn{2}{c}{3D CSG} & \multicolumn{2}{c}{\small ShapeAssembly}\\
    &  CD & Length & CD & Length\\
    \midrule
    PLAD + TTR & $1.13$ &  $\mathbf{15.80}$ & $1.66$ &$\mathbf{8.78}$\\
    \methodshort~ + TTR& $\mathbf{0.83}$ &  $15.96$ & $\mathbf{1.46}$ & $8.80$ \\
    \hline
\end{tabular}
\end{center}
\caption{
Using test-time rewrites with \methodshort~ is superior to using it with PLAD across domains. 
}
\label{table:ttr_on_plad}
\end{table}

\begin{table}[t!]
\begin{center}
  \scalebox{0.80}{
\begin{tabular}{lcccccc}
    \toprule
    & {\small Search} &  {\small only} \doshort & {\small only} \cpshort & {\small only} \cgshort & {\small 1-TTR} & {\small 3-TTR}\\
    \midrule
    IoU & $76.8$ & $81.6$& $76.8$ & $81.98$ & $87.7$ & $90.5$\\
    {\small Length} & $6.81$ & $6.81$ & $6.62$& $14.6$ & $13.52$ & $15.95$  \\
    Time (s) & $1.44$& $1.54$& $0.12$ & $2.81$ & $4.6$ & $14.6$ \\
    \hline
\end{tabular}
}
\end{center}
\caption{
Applying only a single rewriter during test-time, though beneficial, has limited value. Interleaved application of the rewriters which we propose results in a larger improvement. Here, \textit{$n$-TTR} denotes interleaved application of each rewriters $n$ times.
}
\label{table:ttr_sequential}
\end{table}

\section{Conclusion}
\label{sec:conclusion}

We introduced \methodname~(\methodshort~), a paradigm for improving unsupervised training of visual-program inference models with code rewriting.
We implemented a family of code rewriters that generalized across multiple 2D and 3D shape-program domains.
With this family of code rewriters, \methodshort~ learns better VPI networks compared with bootstrap learning methods that ignore rewriters, or use them in a naive fashion.
Beyond this, we demonstrated that our rewriters can be employed in a test-time rewriting (TTR) scheme to improve predictions made by \methodshort~. 
We found that this \methodshort~+ TTR paradigm is able to match or surpass the reconstruction performance of specially designed neural VPI architectures, while maintaining a much more parsimonious program representation. 

In future work, we would like to explore how additional code-rewriting operations could be effectively integrated into our family of rewrites for \methodshort~+ TTR paradigms.
While we find \methodshort~empirically effective for bootstrapped learning, it remains unclear how code rewriting families can best aid RL and end-to-end learning paradigms.
Looking forward, we believe that principled use of code-rewriters is a promising way to guide the search of learning-based VPI models, merging domain-specific preferences with neural guidance, and would be a key component of VPI systems designed for complex, real-world domains.

\section*{Acknowledgment}
\label{sec:acknowledgment}
We would like to thank the anonymous reviewers for their helpful suggestions. This work was funded in parts by NSF award \#1941808 and a Brown University Presidential
Fellowship. Daniel Ritchie is an advisor to Geopipe and owns equity in the company. Geopipe is a start-up that is developing 3D technology to build immersive virtual copies of the real world with applications in various fields, including games and architecture.

{\small
\bibliographystyle{ieee_fullname}
\bibliography{egbib}
}

\end{document}


\title{Improving Unsupervised Visual Program Inference \\ with Code Rewriting Families\\
--Supplementary Material--}

\author{Aditya Ganeshan \qquad R. Kenny Jones \qquad Daniel Ritchie\\
Brown University\\
{\tt\small adityaganeshan@gmail.com} \\
}

\maketitle
\ificcvfinal\thispagestyle{empty}\fi

\noindent
In this document, we provide additional details and experimental results for our proposed method \methodname~. The supplementary material is divided into the following sections:

\begin{itemize}
    \item \textbf{Additional Experiments:} We  provides additional comparison to CSGStump~\cite{CSGSTUMP_ICCV} in section~\ref{subsec-exp-csgstump}. We report our experiments on advanced versions of CSG and ShapeAssembly in Section~\ref{subsec-exp-advanced}, and Section~\ref{subsec-exp-others} reports some additional analysis on \methodshort~.
    \item \textbf{Implementation Details:} In Section~\ref{sec-grammar}, we elaborate on the different Domain Specific Languages (DSLs). Additionally, from section~\ref{subsec-po} - \ref{subsec-cg}, we provide implementation details of the three proposed rewriters.  
    \item \textbf{Qualitative Results:} We present various qualitative comparison between CSGStump~\cite{CSGSTUMP_ICCV}, PLAD~\cite{jones2022PLAD} and our method in section~\ref{sec-qualitative}.
\end{itemize}

\section{Additional Experiments}
\label{sec-exp}

\begin{figure}[t!]
    \centering
   \includegraphics[width=1.0\linewidth]{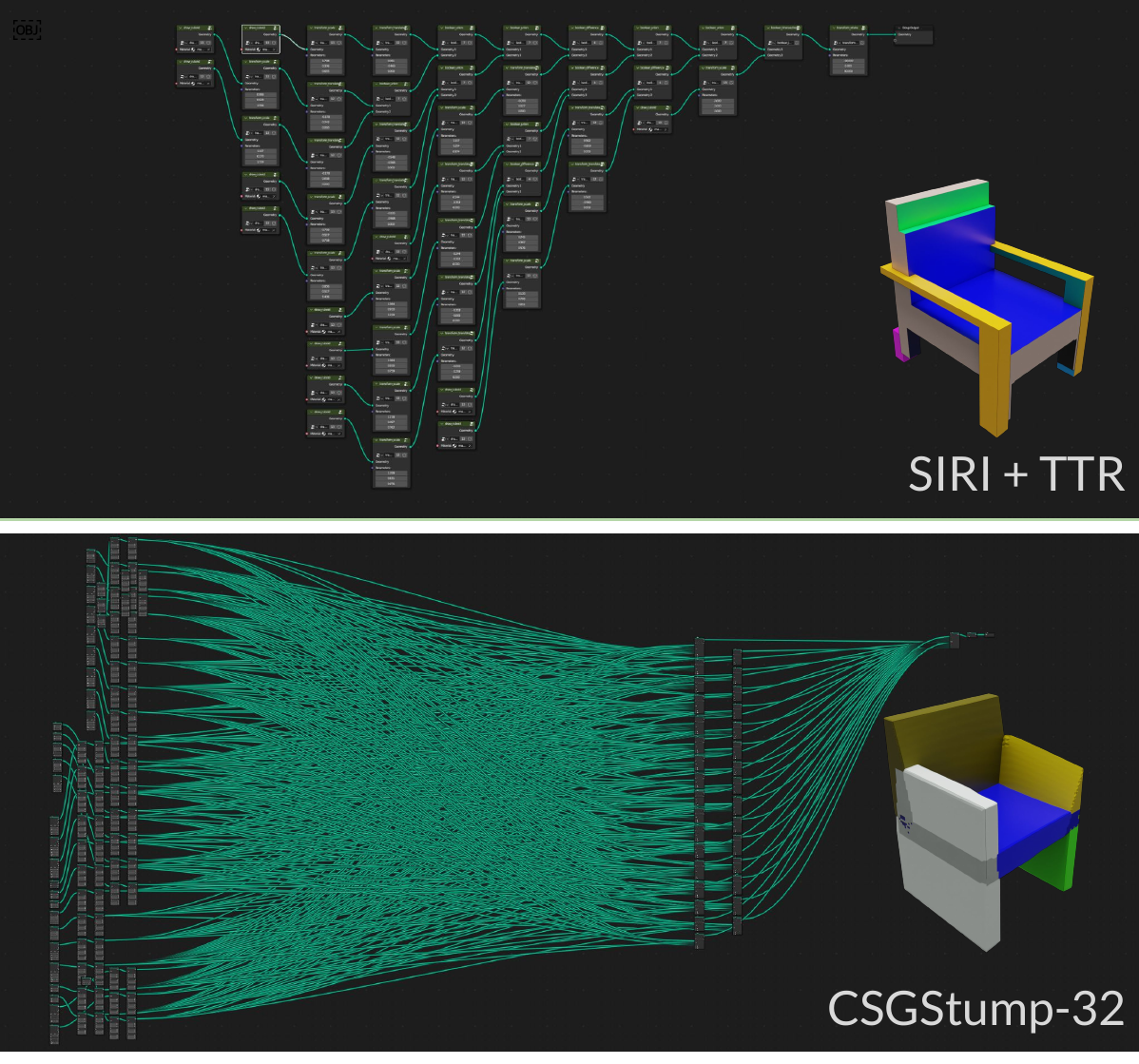}
    \caption{
    We show the 3D CSG programs inferred by different methods as Geometry Node trees in Blender~\cite{blender}. Programs inferred by even the smallest CSGStump~\cite{CSGSTUMP_ICCV} model (with 32 primitives and 32 intersection nodes) is highly complicated. In comparison, \methodshort~ programs are parsimonious and hence easier to edit.
    }
    \label{fig:blender_graphs} 
    
\end{figure}

\subsection{Comparison to CSG-Stump}
\label{subsec-exp-csgstump}

\noindent
\textbf{Program Complexity:} In Figure~\ref{fig:blender_graphs}, we visualize the programs inferred by CSG-Stump 32 and \methodshort~ as Geometry Node graphs in Blender~\cite{blender}, a popular tool for procedural modelling. We can see that CSG-Stump 32, despite being the smallest of the models (vs CSG-Stump 256), still produces highly complex graphs with a myriad of nodes and links. 
Editing such programs to create shape variations is arduous. Additionally, the program is hard to interpret due to its highly connected nature. Such complex program graphs lose a fundamental advantage of the programmatic representation: interpretability and editability.

\begin{table*}[t]
\begin{center}
\begin{tabular}{llcccccc}
\toprule

 TTR & Model& {\small IoU-32 ($\uparrow$)}& {\small IoU-64 ($\uparrow$)} & {\small mean CD ($\downarrow$)} &  {\small med. CD ($\downarrow$)}&{\small Length ($\downarrow$)}&{\small Time ($\downarrow$)}\\
\midrule
  $\emptyset$ & {\small CSG-Stump 32} & $49.88$ & $39.83$ & $1.88$ & $1.58$ & $98.94$ &$0.05$ \\
$3$ & {\small CSG-Stump 32} & $73.83$ & $54.34$ & $2.56$ & $1.03$ & $63.58$ &$133.88$ \\
\midrule
 $\emptyset$ & {\small \methodshort~} & $76.77$ & $54.61$ &$1.12$ & $0.69$ & $6.81$& $1.44$ \\
 $3$ & {\small \methodshort~} & $90.59$ & $60.60$ & $0.83$ & $0.51$ &$15.96$  &$14.60$\\
\bottomrule
\end{tabular}
\end{center}
\caption{
We perform test time rewriting on programs inferred by CSG-Stump 32~\cite{CSGSTUMP_ICCV} with our proposed rewriters. While it improves the reconstruction accuracy and decreases the program length, it still under performs \methodshort~. Additionally, applying TTR to CSG-Stump 32 inferred programs takes a order of magnitude more time than applying TTR to \methodshort~ inferred programs. 
}
\label{table:csgstump-main}

\end{table*}
 
\noindent
\textbf{Test-time rewriting with CSG-Stump:} 
Our work proposes three rewriters which can improved bootstrapped learning as well as perform test time rewriting (TTR). An important question is whether TTR can 
improve programs inferred by domain specific networks such as CSG-Stump~\cite{CSGSTUMP_ICCV}. To answer this question, we perform test time rewriting with our rewriters on programs inferred by CSG-Stump. We report the results in Table~\ref{table:csgstump-main}. Note that to limit the optimization time, we apply our rewriters on CSG-Stump 32 model (a model with 32 primitives, and 32 intersection nodes). 

First, we notice that TTR indeed improves the reconstruction quality (as measured by IoU and median CD) as well as the program length for CSG-Stump as well. However, we do note that mean CD spikes up due to some poorly optimized programs. Second, we note that due to the large programs inferred by CSG-Stump (even for the smallest model), TTR takes an order of magnitude more time than it takes with \methodshort~ inferred programs. Both of these trends indicate that producing sparse programs from start (via bootstrapped methods) may be preferable over producing over-parameterized programs (via domain specific networks).

\begin{table}[t!]
\begin{center}
  \scalebox{1.00}{
    \begin{tabular}{lccc}
    \toprule
     & \multirow{1}{*}{\small IoU ($\uparrow$)}  &  {\small  CD ($\downarrow$)} &  {\small Length ($\downarrow$)}  \\
    \midrule
    \methodshort~ & $\mathbf{78.63}$ & $\mathbf{1.13}$ & $6.81$\\
    \midrule
    No Sparse & $77.94$ & $1.28$ & $6.98$\\
    Single Rewrite Queue & $77.45$ & $1.34$ & $\mathbf{6.15}$\\
    Single Queue & $77.63$ & $1.25$ & $6.53$\\
    \midrule
    \pplusrshort~ &  $75.91$ & $1.29$ & $8.98$\\
    \bottomrule
    \end{tabular}
    }
\end{center}
    \caption{
    We compare variants of \methodshort~ to validate its design. \methodshort~ outperforms the baselines, and shows that \textit{sparse} rewrite usage, as well as \textit{careful} rewrite injection (via source-separated queues) are essential to integrate rewriters into bootstrapped learning processes.
    }
    \label{table:siri-ablation}
    
\end{table}

\subsection{Ablation}
\label{subsec-exp-ablation}

We now provide an ablative analysis of our method to verify the importance of each component. For consistency, we perform all ablations on the 3D CSG language and report metrics on the validation set.

\noindent
\textbf{\methodshort~ ablation:} We perform a subtractive analysis on \methodshort~ by changing/removing individual components of the method. We compare \methodshort~ to three alternatives:

\begin{packed_enumerate}
    \item \textbf{No Sparse:} Instead of applying to rewriters to a subset of programs, they are applied to all programs. 
    \item \textbf{Single Rewrite Queue:} Instead of storing separate queues for each rewriter via the source mapping $S_{PO}, S_{CP}, S_{CG}$, we store a single queue $S_R$ shared between all the rewriters.
    \item \textbf{Single Queue:} We remove the separate sources $S_{NS}, S_{PO}, S_{CP}, S_{CG}$, and simply store the top-k ($k=3$) programs for each input shape $x$.
\end{packed_enumerate}

We report the results in Table~\ref{table:siri-ablation}. \methodshort~ surpasses all the other ablations. At the same time, all the ablations are able to surpass the naive rewriter integration \pplusrshort, albeit with a smaller margin. This shows that \textit{sparse} rewrite usage, as well as \textit{careful} rewrite injection (via source-separated queues) help integrating the rewriters into bootstrapped learning processes.

\begin{table}[t!]
\begin{center}
    \begin{tabular}{lccc}
    \toprule
     & \multirow{1}{*}{\small IoU ($\uparrow$)}  &  {\small  CD ($\downarrow$)} &  {\small Length ($\downarrow$)}  \\
    \midrule
    \methodshort~ & $\mathbf{78.63}$ & $\mathbf{1.13}$ & $6.81$\\
    \midrule
    no \doshort & $77.80$ & $1.27$ & $ 6.10 $\\
    no \cpshort & $77.31$ & $1.21$ & $6.18 $ \\
    no \cgshort & $76.21$ & $1.27$ & $\mathbf{5.58}$ \\
    \midrule
    PLAD & $76.21$ & $1.43$ & $6.39$ \\
    \bottomrule
    \end{tabular}
\end{center}
    \caption{
    We report reconstruction accuracy as well as code-quality on 3D CSG+ with each rewriters individually removed. Using all the three rewriters (top-row) yields the best reconstruction and code-quality. 
    Using no rewriters is equivalent to PLAD (bottom-row).  
    }
    \label{table:rewriter-ablation}
\end{table}

\noindent
\textbf{Rewriter Ablation:} Next, we test whether using all the three proposed rewriters \doshort, \cpshort, \cgshort is essential. Table~\ref{table:rewriter-ablation} compares \methodshort~ to models trained with one rewriter removed each. We see that using all the three rewriters improves the performance.

\subsection{Advanced Languages}
\label{subsec-exp-advanced}

To take a step towards supporting languages closer to those used in real-world scenarios, we extend 3D CSG and ShapeAssembly to contain more complex features (described in detail in Section~\ref{sec-grammar}), and perform additional experiments on these languages. Note that while the extensions are simple under our general framing, domain-specific networks such as CSG-Stump~\cite{CSGSTUMP_ICCV} and UCSG-Net~\cite{kania2020ucsgnet} require non-trivial changes in their architecture to support such extensions.

\begin{table*}[t!]
\begin{center}
  \scalebox{1.0}{
\begin{tabular}{lcccccc}
    \toprule
    & \multicolumn{3}{c}{3D CSG+} & \multicolumn{3}{c}{ShapeAssembly+} \\
    &{\small IoU ($\uparrow$)} & {\small CD ($\downarrow$)}& {\small Length ($\downarrow$)} &{\small IoU ($\uparrow$)} & {\small CD ($\downarrow$)}& {\small Length ($\downarrow$)} \\
    \midrule
    PLAD   & $70.51$ & $2.35$ & $\mathbf{9.09}$& $61.22$ & $2.63$ & $\mathbf{8.51}$  \\
    \pplusrshort  & $68.25$& $2.45$ & $10.57$ &  $58.82$ & $2.74$ & $10.31$ \\
    \midrule
    \methodshort~ & $\mathbf{71.48}$ & $\mathbf{2.17}$& $9.99$ & $\mathbf{63.91}$ & $\mathbf{2.41}$ & $9.24$ \\
    \bottomrule
\end{tabular}
}
\end{center}

\caption{
We report the Test-set performance across advanced versions of 3D CSG (3D CSG+) and ShapeAssembly (ShapeAssembly+). As a result of their higher complexity, we see performance on advanced languages is lower than performance on simple languages. However, that trend observed in the main draft is seen here as well - naively integrating the rewriters into PLAD (\pplusrshort) can deteriorate the model's performance, and \methodshort~ consistently outperforms both the baselines.
}
\label{table:advanced-main}

\end{table*}

\begin{table}[t!]
\small
\begin{center}
  \scalebox{1.0}{
\begin{tabular}{lcccccc}
    \toprule
   \multirow{2}{*}{\small TTR} &  \multicolumn{3}{c}{3D CSG+} & \multicolumn{3}{c}{\small ShapeAssembly+}\\
    & IoU & CD & Length & IoU & CD & Length\\
    \midrule
    PLAD & $78.5$ &$1.68$ &  $\mathbf{15.9}$ & $71.82$ & $2.17$ & $\mathbf{11.35}$\\
    \methodshort~& $\mathbf{82.8}$ &$\mathbf{1.48}$ &  $20.3$ & $\mathbf{74.65}$ & $\mathbf{1.92}$ & $11.47$ \\
    \hline
\end{tabular}
}
\end{center}
\caption{
We apply test time rewriting for models trained on the advanced languages. Reaffirming the results in the main draft, we see that using test-time rewrites with \methodshort~ remains superior to using it with PLAD.
}
\label{table:advanced-ttr}

\end{table}

\noindent
\textbf{Bootstrapped Learning:} We train the unsupervised VPI network with PLAD, \pplusrshort and \methodshort~. We report the results in Table~\ref{table:advanced-main}. On both the advanced languages, \methodshort~ outperforms the baselines. More importantly, a naive integration of the rewrites (i.e. \pplusrshort) is detrimental on both the languages. This experiment emphasizes the complexity of integrating rewrite mechanisms into learning frameworks. By using the rewriters sparsely and carefully injecting rewritten programs into the training data, \methodshort~ is able to improve bootstrapped learning through the rewriters.   

\noindent
\textbf{Test Time Rewriting:} We perform test time rewriting (TTR) on both 3D CSG+ and ShapeAssembly+ domains with models trained with PLAD and \methodshort~. Our results are tabulated in Table~\ref{table:advanced-ttr}. Similar to the results in the main draft, test time rewriting is more effective on programs inferred by \methodshort~ than by PLAD. 

\subsection{Other Experiments}
\label{subsec-exp-others}

\begin{figure}[t]
    \centering
   \includegraphics[width=1.0\linewidth]{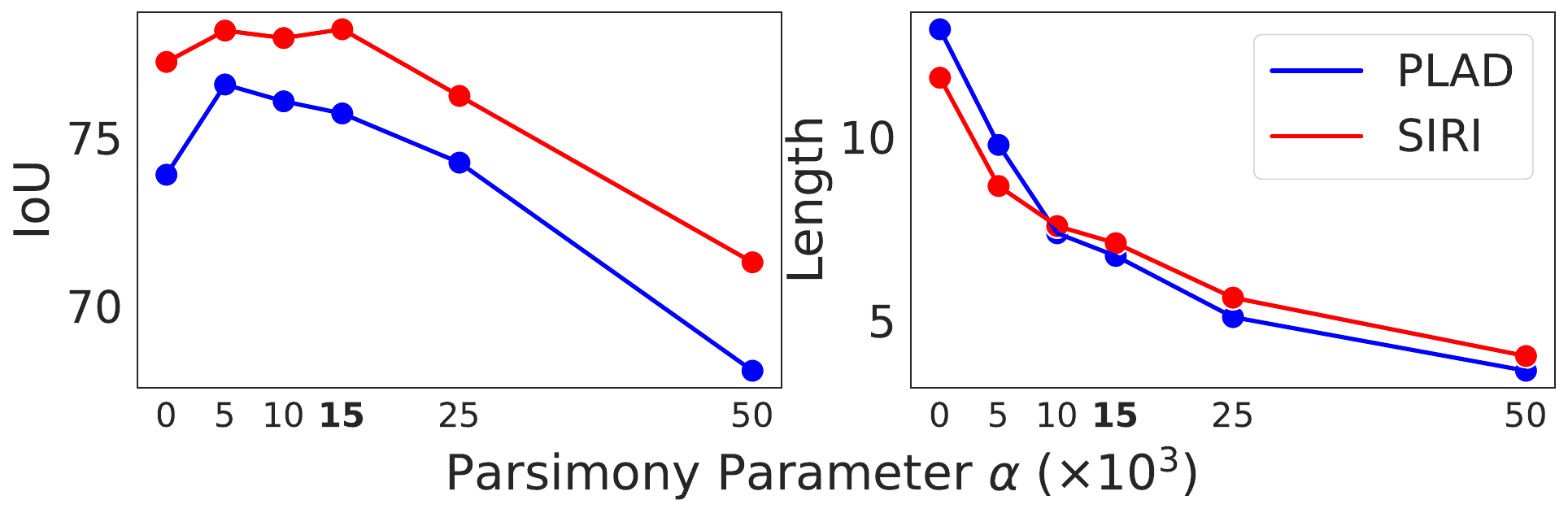}
    \caption{\small 
    We compare PLAD and \methodshort~ for different values of $\alpha$ (X-axis), measuring the IoU (\textit{left}), and program-length (\textit{right}) on the validation set of the 3D CSG domain.
    \methodshort~ consistently dominates PLAD, and setting $\alpha=0$ harms both reconstruction accuracy and conciseness. 
    }
    \label{fig:alpha} 
\end{figure}

\noindent
\textbf{Sensitivity to $\alpha$:} A key ingredient in the objective $\mathcal{O}$ is the $\alpha$ factor balancing the weightage between reconstruction accuracy and program length. In figure~\ref{fig:alpha}, we compare the validation-set performance of PLAD and \methodshort~ on 3D CSG for different values of the alpha parameter. We note that across the different settings, \methodshort~ consistently outperforms PLAD in reconstruction accuracy (IoU).

\begin{figure*}[t!]
    \centering
   \includegraphics[width=1.0\linewidth]{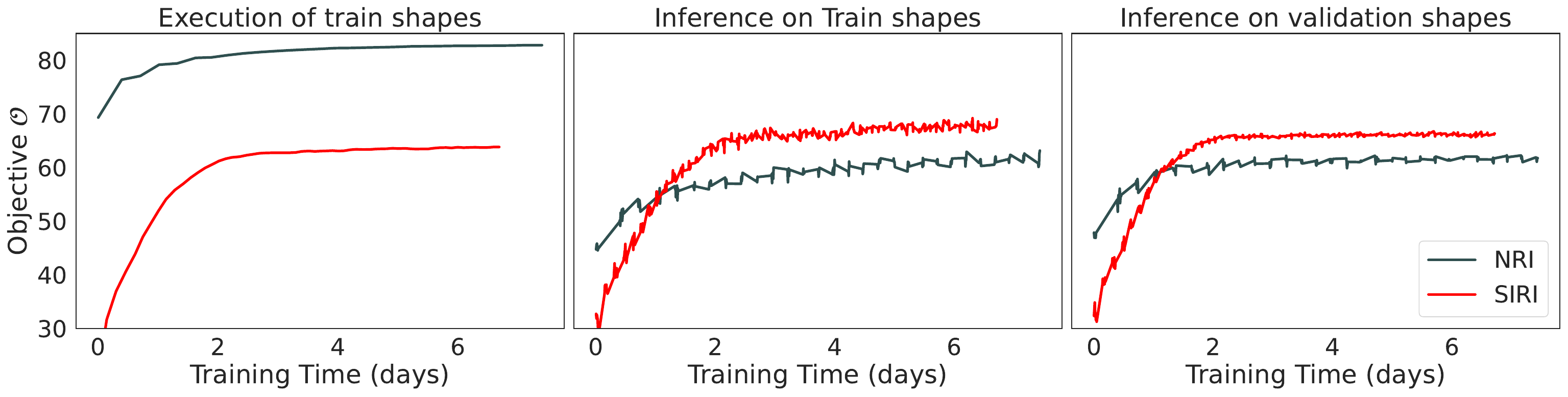}
    \caption{
    To understand why \pplusrshort fails, we analyse the training process for \pplusrshort trained and \methodshort~ trained models. We show the (a) execution of train shapes (b) inference on train shapes and (c) inference on validation shapes. \pplusrshort does not overfit to the training data, and while its training data is of high quality (i.e. their execution yields high value of objective $\mathcal{O}$), its train and validation inference performance remain low (cf. section~\ref{subsec-exp-others} for more details).
    }
    \label{fig:nri_analysis} 
\end{figure*}

\noindent
\textbf{Understanding \pplusrshort failure:} Why does \pplusrshort fail? To answer this question, we study additional training statistics, namely the trained network's train-set performance, val-set performance, and quality (as measured by our objective $\mathcal{O}$) of the training data created via the \search and \rewrite phases. We show the corresponding graphs in Figure~\ref{fig:nri_analysis}. 

One may assume that \pplusrshort performs poorly due to overfitting to the training data. However, we see that this is not the case - inference performance on both the training-set as well as validation-set for \pplusrshort is lower than \methodshort~. As PLAD-finetuning performs early-stopping, with weight-reloading before \search phase, it avoids overfitting to the training data. 
Another hypothesis may be that the programs found for the training shapes might be bad matches. However, we see that this is also not the case from the `Execution of train shapes' plot. As \pplusrshort only trains on the ``best programs'' seen thus far, the training data quality, as measured by the objective $\mathcal{O}$, is significantly higher than \methodshort~. However, despite the higher reconstruction matches against their target shapes, the model trained with \pplusrshort does not generalize from these training programs and fails to perform well on the validation-set, resulting in performance stagnation for \pplusrshort. In contrast, \methodshort~ is able to perform well on the validation set, despite training on programs which have lower reconstruction matches versus their target shapes. 
This indicates that for bootstrapped learning, simply maximizing the reconstruction accuracy of training programs may not be sufficient: instead it is necessary to find programs for shapes in the training set that both (i) produce good reconstructions, and (ii) help the inference network to learn policies that generalize to held-out shapes from the same-distribution (validation set).
We find that the excessive use of rewriters in \pplusrshort violates assumption (ii), but we hope to explore this phenomena more in future work.

\section{Evaluation details}
\label{sec-exp-eval-details}
\noindent
To infer programs from the networks learned via bootstrapped learning processes (PLAD, \pplusrshort, \methodshort~), we perform neurally guided-beam search with a beam-size of $10$. For each input shape, this search generates a candidate pool of programs, from which, the objective maximizing program is selected as the inferred program. This approach is the same as the one used in PLAD~\cite{jones2022PLAD}. For CSG-Stump, a forward pass through the network produces a single probabilistic program. The inferred program is then generated by simply replacing each probabilistic parameter with the modal value of its distribution (for instance, the argmax of each categorical distributions).

Given a set of input shapes, we measure the quality of inferred programs along two axis: a) reconstruction accuracy, to measure how well the inferred program's execution reconstructs the input shape, and b) program parsimony, to measure how concise the inferred programs are (motivated by the Occam's razor principle). 

\noindent
\textbf{Reconstruction metrics:}
    The execution of programs results in a $\mathcal{R}^n$ grid of occupancy values (where $n=2$ for 2D and $n=3$ for 3D). As the input is also in the form of a $\mathcal{R}^n$ grid of occupancy values, we can directly measure the Intersection over Union between the inferred occupancy and ground-truth occupancy by comparing them. For measuring Chamfer Distance, on 2D data we follow CSG-Net~\cite{CSGNet}, and on 3D data we follow CSG-Stump~\cite{CSGSTUMP_ICCV}. 

\noindent
\textbf{Program Parsimony:} Program length is measured as the number of statements/commands in a program. Note that this differs from the number of discrete tokens required to define a command/statement (for instance, a Cuboid command in 3D CSG requires 7 tokens, 1 to specify command type, 3 to specify the 3D position, and 3 to specify the 3D scale). 

\section{DSL Grammars}
\label{sec-grammar}

\subsection{Simple DSLs}

\noindent
\textbf{2D CSG:} For 2D CSG, each primitive is instantiated with $5$ parameters, specifying its 2D position, 2D scale and rotation. 
We specify the grammar as follows:
\begin{align*}
    S &\rightarrow E; \\
    E &\rightarrow BEE\ |\ D(F, F, F, F, F); \\
    B &\rightarrow intersect\ |\ union\ |\ subtract;\\
    D &\rightarrow rectangle\ |\ ellipse; \\
    F &\rightarrow (-1, 1);
\end{align*}

In the $D(F, F, F, F, F)$ command, the first two real numbers $F$ specify the translation, the next two specify scaling, and the last specifies rotation. $F$ is mapped to different ranges for these different operations. For translation, $F$ remains as is, for scaling $F$ is linearly mapped from $(-1, 1)$ to $(0.01, 2.01)$, for rotation, we linearly map $F$ from $(-1, 1)$ to $(-\pi, \pi)$.

\noindent
\textbf{3D CSG: }3D CSG matches the language used in PLAD~\cite{jones2022PLAD}.  For 3D CSG, each primitive is defined by $6$ parameters, specifying its 3D position, and 3D scale (no rotation). 
The range mapping of real number language tokens $F$ for the different operations followed for 2D CSG is applied here as well.
We specify the grammar as follows:
\begin{align*}
    S &\rightarrow E; \\
    E &\rightarrow BEE\ |\ D(F, F, F, F, F, F); \\
    B &\rightarrow intersect\ |\ union\ |\ subtract;\\
    D &\rightarrow cuboid\ |\ ellipsoid\ ;\\ 
    F &\rightarrow [-1, 1];
\end{align*}

\noindent
\textbf{ShapeAssembly:} ShapeAssembly creates structures by instantiating cuboids, and attaching them to one another. We utilize the version of ShapeAssembly described in PLAD~\cite{jones2021shapeMOD} with a slight modification. Specifically, in our version $\texttt{attach}$ and $\texttt{squeeze}$ commands are defined for 3D points rather than 2D surface points. This makes our version of ShapeAssembly closer to the original defined in ShapeAssembly~\cite{ShapeAssembly}. 
Following PLAD, we restrict the bounding box to always have $l=1, w=1$, and remove the axis-alignment flag from cuboid instantiation. Furthermore, for all sub-programs we restrict the bounding box to always have $h=1$ as well. 
We specify the grammar for ShapeAssembly as the following:

\begin{align*}
&Start  \rightarrow  BBlock; CBlock; ABlock; SBlock; \\
&BBlock \rightarrow \text{bbox} = \texttt{Cuboid}(l, h, w)  \\
&CBlock \rightarrow c_n = \texttt{Cuboid}(l, w, h) ;  CBlock\: |\: None \\
&ABlock \rightarrow A ; ABlock\: |\:  S ; ABlock \: |\: None \\
&SBlock \rightarrow R ; SBlock\: |\:  T ; SBlock \: |\: None \\
&A \rightarrow \texttt{attach}(c_{n_1}, x_1, y_1, z_1, x_2, y_2, z_2) \\
&S \rightarrow \texttt{squeeze}(c_{n_1}, c_{n_2}, f, u, v) \\
&R \rightarrow \texttt{reflect}(\text{axis}) \\
&T \rightarrow \texttt{translate}(\text{axis}, m, d) \\
&f \rightarrow right\: |\: left \:|\: top\: |\: bot\: |\: front\: |\: back \\
&\text{axis} \rightarrow X\: |\: Y \:|\: Z\: \\
&l, h, w \in \mathbb{R}^+ \\
&x, y, z, u, v, d \in [0,1]^2 \\
&n, m \in \mathbb{Z}^+ \\
\end{align*}
\subsection{Advanced DSLs}

To ease learning, past approaches have used \textit{simplified} versions of visual languages as described in the previous section. However, real world visual programs offer advanced operations, e.g. real world CSG programs contain hierarchical transformations (rather than primitive level transformations). As a step towards such languages, we also test our method on \textit{advanced}  3D CSG (\textbf{3D CSG+}) and ShapeAssembly (\textbf{ShapeAssembly+}).

\noindent
\textbf{3D CSG+:} we introduce two important classes of commands - \textit{hierarchical transformations}, transformations that can be applied to compound shapes, and \textit{axis-aligned reflection}. Further, to reduce parameter redundancy, we instantiate primitives without any parameters, i.e., they are instantiated in a canonical form (origin centered, and unit scale with no rotation). The grammar is defined as follows:
\begin{align*}
    S &\rightarrow E; \\
    E &\rightarrow BEE\ |\ TE \ |\ D; \\
    B &\rightarrow intersect\ |\ union\ |\ subtract;\\
    D &\rightarrow cuboid\ |\ ellipsoid\ |\ cylinder;\\
    T &\rightarrow translate(F, F, F)\ |\ scale(F, F, F)\ |\ \\
    &\quad \ \  rotate(F, F, F)\ |\ R; \\
    R &\rightarrow reflect(X)\ |\ reflect(Y)\ |\ reflect(Z);\\ 
    F &\rightarrow [-1, 1];
\end{align*}

\noindent
\textbf{ShapeAssembly+:} We extend the simple ShapeAssembly described previously by allowing hierarchical composition of programs. Specifically, our version allows hierarchical programs, where cuboids in the root program can act as the bounding box for their own ShapeAssembly programs.  
ShapeAssembly+ follows the same grammar as ShapeAssembly with the following change:
\begin{align*}
&CBlock \rightarrow c_n = \texttt{Cuboid}(l, w, h, \mathbf{sp}) ;  CBlock\: |\: None \\
& \mathbf{sp}  \in \mathbb{Z}^+ \\
\end{align*} 
During cuboid instantiation, parameter $\mathbf{sp}$ specifies whether the cuboid is empty ($sp =0$) or contains the $sp$-th ShapeAssembly program (all the ShapeAssembly programs are decoded by the inference network).

\section{Code Rewriters}
Our rewriters are formulated to be generally applicable across visual programming languages. 
In this section, we outline the requirements to apply each rewriter, along with implementation details and a walkthrough example.

\subsection{Parameter Optimization (PO)}
\label{subsec-po}
The \doname(\doshort) rewriter aims to improve the continuous parameters of a given program while keeping its discrete structural parameters fixed. Furthermore, it aims to use first-order gradient based optimization to update these parameters. 
Towards this end, \doshort requires that the programs derived from the language grammar must be continuous and piecewise differentiable with respect to $\ge 1$ program parameters $\phi$. 

Additionally, as \doshort performs gradient based optimization, it requires a  \textit{partially} differentiable executor for the language (differentiable w.r.t. $\phi$ at least). 
By chaining a differentiable reconstruction measure $\mathcal{R}$ (such as $L_2$-loss w.r.t. a occupancy grid) to the program's differentiable execution, we can then optimize the program parameters $\phi$ to improve the reconstruction measure $\mathcal{R}$. 

Most shape languages instantiate parameterized primitives and combine them with different combinators to create shapes. When languages can map the program parameters $p$ to primitive parameters in a piecewise differentiable manner (as in ShapeAssembly~\cite{ShapeAssembly}, and CSG), we can yield a \textit{partially} differentiable executor $D$ for the language with the procedure outlined in the paper (cf. Section 4.1). We revisit the procedure here with more details.

\begin{figure}[t!]
    \centering
   \includegraphics[width=1.0\linewidth]{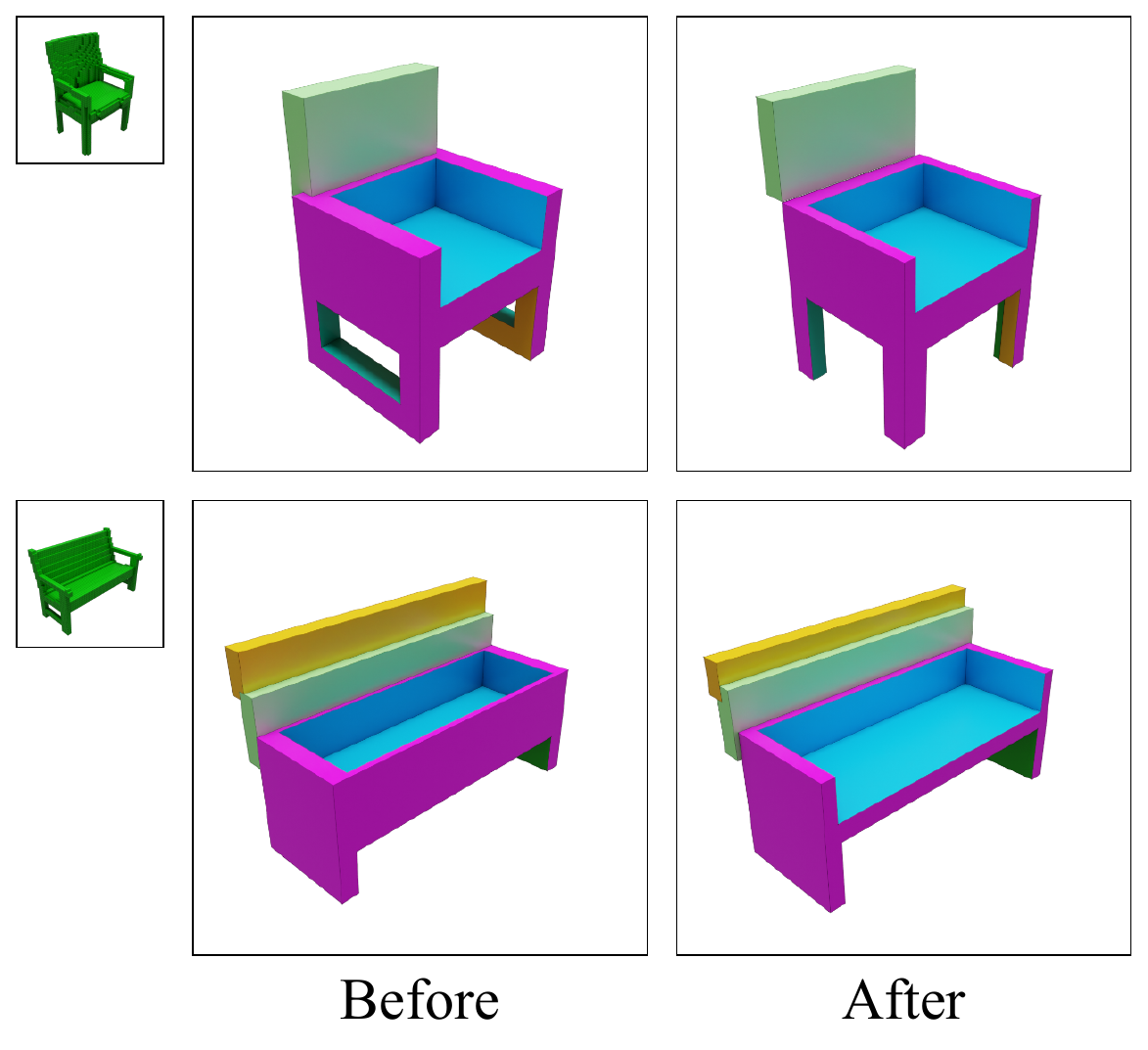}
    \caption{
    We show the \doshort rewriter applied to 3D CSG programs. 
    }
    \label{fig:po_examples} 
\end{figure}

$D(z)$ maps the execution of program to an equivalent implicit sign-distance function. Given a program $z_{\phi}$, we use the programs executor $E$ to map program parameters $\phi$ to parameters (position, scale, rotation) of implicit functions representing simple geometric primitives (cuboids, spheres etc.). For CSG, the mapping is 1-to-1, but for other languages such as ShapeAssembly, these parameters are derived from the program's execution. 
Then, we apply boolean combinators of the parameterized primitives to obtain the program's \emph{implicit equivalent}. 
Armed with the program's implicit equivalent, we uniformly sample points $t \in \mathbb{R}^n$ and convert the signed distance at the points into ``soft'' occupancy values to yield a differentiable execution of the program.

\noindent
\textbf{Optimization:} From the program's \textit{implicit equivalent}, the soft-occupancy values $O(z)$ are derived as follows:
\begin{equation}
    O(z) = \sigma( - tanh ( sdf(z) \times \alpha) \times \alpha),
\end{equation}
where $\sigma$ is the sigmoid function, and $\alpha$ is a scalar value. 
We run our optimization procedure for $250$ steps with the Adam~\cite{Adam} optimizer (learning rate is set as $0.01$), and we scale $\alpha$ logarithmically from $log(3)$ to $log(10)$. Finally, to ensure that the parameter values remain within range, we perform our optimization on $\theta = tanh^{-1}(\phi / s)$,  where $s$ is the range of each parameter, instead of $\phi$ directly. Figure~\ref{fig:po_examples} presents examples of the \doshort rewriting procedure in action.

\begin{figure}[t!]
    \centering
   \includegraphics[width=1.0\linewidth]{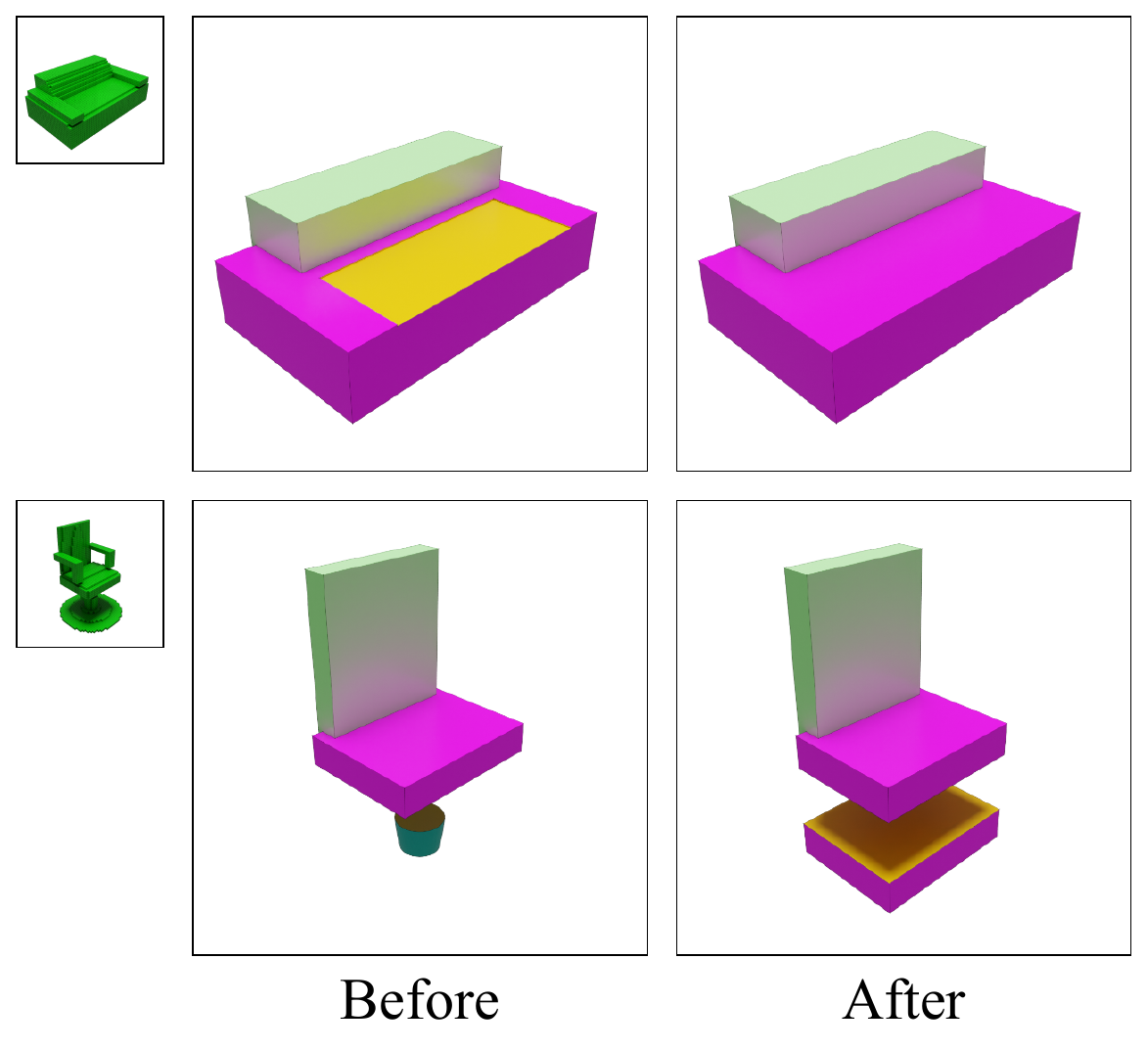}
    \caption{
    We show the \cpshort rewriter applied to 3D CSG programs. Note that in the second figure, \cpshort identifies a sub-tree of the input program which in fact obtains higher reconstruction accuracy w.r.t. the target shape. \
    }
    \label{fig:cp_examples} 
\end{figure}

\begin{figure*}[t!]
    \centering
   \includegraphics[width=0.95\linewidth]{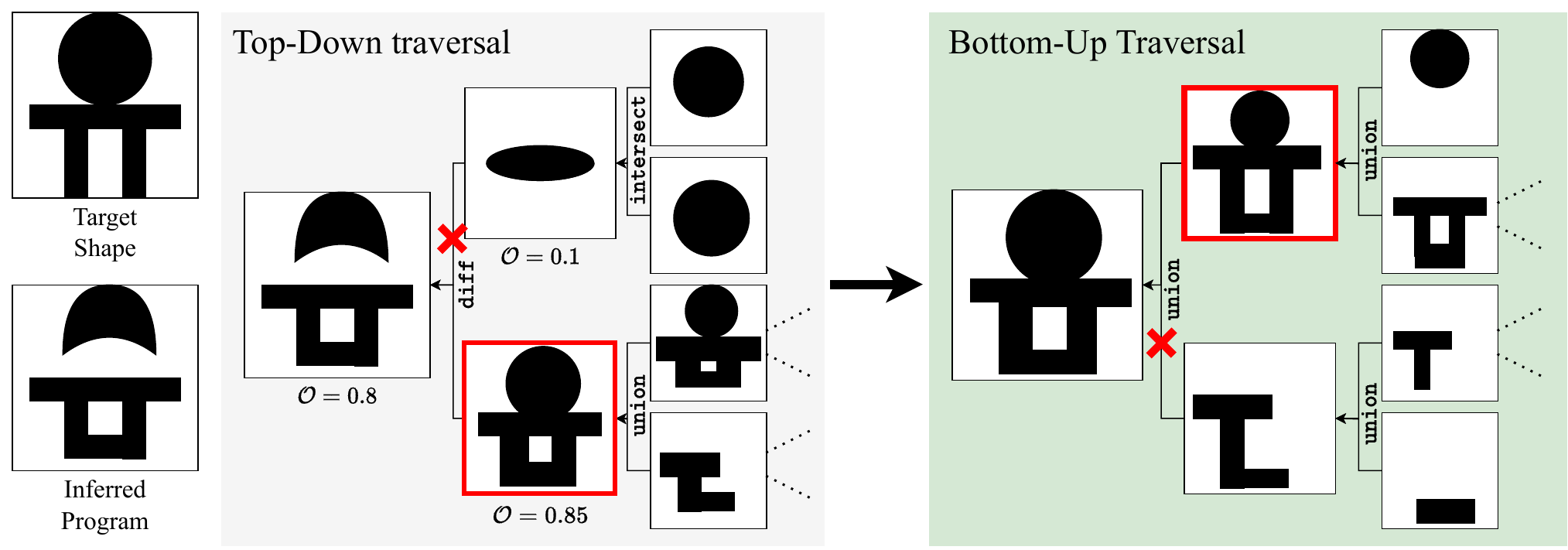}
    \caption{
    We show the two traversals of \cpshort rewriter on 2D CSG on a manually created example. The Top-Down traversal identifies the objective $\mathcal{O}$ maximizing root node, and Bottom-Up traversal prunes extraneous nodes of the sub-tree starting from objective maximizing node. In this example, during the bottom-up traversal, as the top node matches its parent node, we remove its sibling node and simplify the expression.
    }
    \label{fig:td_bu} 
\end{figure*}

\subsection{Code Pruning (CP)}
\label{subsec-cp}

Given a shape $x$ and program $z$, \cpshort rewrites $z$ s.t. $z^R \sim \argmax_{\Omega^{CP}} \mathcal{O}(x, z)$, where $\Omega^{CP}(z) = \{\tilde z | \tilde z \subseteq z \}$ represents the set of all valid sub-programs of $z$, i.e. \cpshort aims to identify and remove program fragments that negatively contribute to our objective $\mathcal{O}$.  
In Figure~\ref{fig:cp_examples}, we show examples of \cpshort rewriter in action and in Algorithm~\ref{alg-cp} we present a high-level overview of \cpshort.
We first describe how to apply \cpshort to a simply-typed lambda expression. We then describe how we map different languages, such as CSG and ShapeAssembly to such expressions.

Given a simply-typed lambda expression, we perform \cpshort via two rewrite passes, namely a bottom-up and a top-down pass to approximate $z^R$. 
First, we create a directed acyclic expression-tree comprised of expression-nodes and edges connecting them. Each expression node stores the execution of the sub-expression formed with that node as the root. Now our goal is to identify nodes which can be pruned from the graph. First, we perform a top-down traversal of the graph, and evaluate the objective $\mathcal{O}$ w.r.t. the input shape $x$ at each node. We identify the node with the highest score, and mark it as the root node.  
Next, during the bottom up traversal, we identify and prune nodes which are extraneous. 
For CSG language, we detect such extraneous nodes by comparing them to their parent node -  if the parent node's execution exactly matches a child node's execution, all the sibling nodes are extraneous and can be removed. We also mark nodes as extraneous if the node's execution is empty (i.e. the node's subexpression evaluates to null). For ShapeAssembly, nodes whose execution does not overlap with the target shape $x$ are marked as extraneous (as ShapeAssembly is a purely additive language).
We present an example of the two traversals in \cpshort rewriting procedure in Figure~\ref{fig:td_bu}.

\begin{algorithm}[t]
\caption{\cpname}\label{alg-cp}
\begin{flushleft}
\textbf{Input:} Set of programs $Z_x$ \& Executor $E$.\\
\textbf{Hyper-parameters:} no. programs to rewrite $n$\\
\textbf{Output:} Rewritten programs $Z_R$\\
\vspace{-0.5em}
\hrulefill
\vspace{-0.5em}
\begin{algorithmic}[1]
\STATE \textbf{for} $z_x$ in random\_sample($Z_x$, $n$):
\STATE $\quad$ {\color{gray} \# top down pass}
\STATE $\quad$ $z_{r}$ = $z_x$
\STATE $\quad$ \textbf{for} \textit{subtr\_node} in extract\_subtree\_nodes($z_x$):
\STATE $\quad \quad$ \textbf{if} $\mathcal{O}(z_{subtr\_node}, x) \ge \mathcal{O}(z_{r}, x)$:
\STATE $\quad \quad \quad$ $z_{r} = z_{subtr\_node}$
\STATE $\quad$ {\color{gray} \# bottom up pass}
\STATE $\quad$ \textbf{for} \textit{subtr\_node} in extract\_subtree\_nodes($z_r$):
\STATE $\quad \quad$ \textbf{if} removable($z_{r}$, \textit{subtr\_node}):
\STATE $\quad \quad \quad$ $z_{r}$ = remove($z_{r}$, \textit{subtr\_node})
\STATE $\quad $ \textbf{if} $\mathcal{O}(z_r, x) \ge \mathcal{O}(z_{x}, x)$:
\STATE $\quad \quad$ $Z_{R}$.insert($z_{r}$)
\end{algorithmic}
\end{flushleft}
\end{algorithm}

\begin{figure*}[t!]
    \centering
   \includegraphics[width=1.0\linewidth]{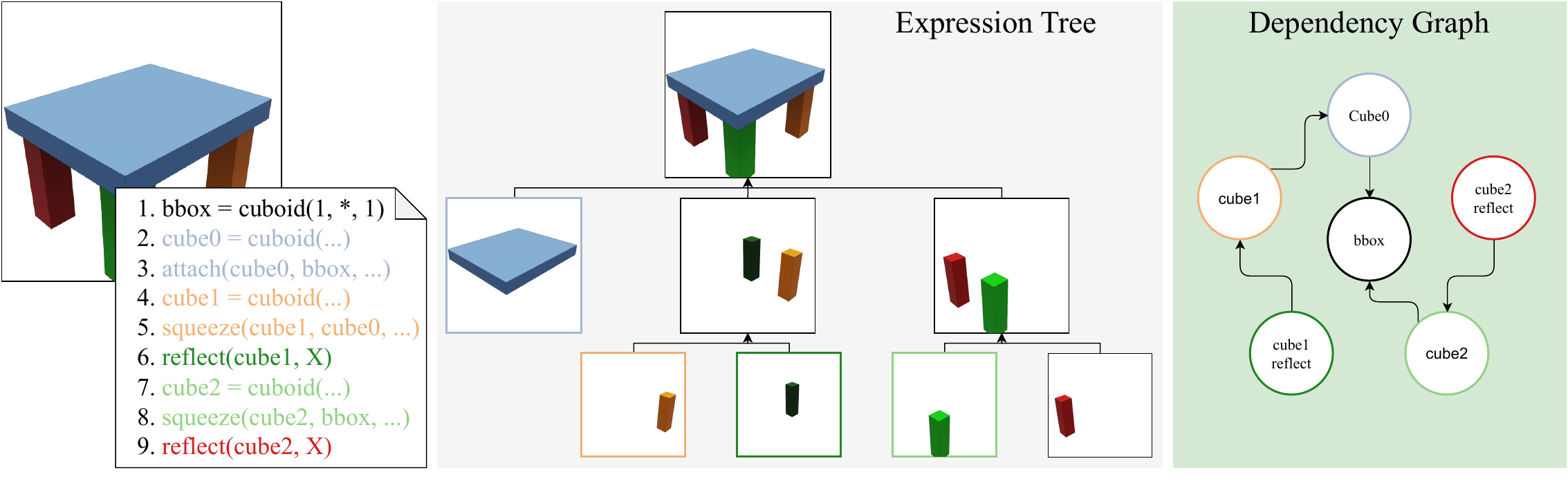}
    \caption{
    We show the mapping of a ShapeAssembly program to (a) its implicit equivalent expression tree and (b) its dependency graph. The \cpshort rewriter prunes extraneous nodes from the expression tree while also ensuring pruned nodes have no incoming dependency. For the illustrated example, only the parts created by reflection operators are prune-able, as they are the only leaves in the dependency graph.
    }
    \label{fig:sa_cp} 
\end{figure*}

As stated earlier, to apply \cpshort, we map the program to a simply typed lambda expression, which is then used to construct an expression tree. For CSG, the program itself is such an expression. For ShapeAssembly, we utilize the program's implicit equivalent (as defined in Section~\ref{subsec-po}) to construct the expression. However, since ShapeAssembly is an imperative language, nodes can have interdependency which restricts their removal. Therefore, we additionally extract a partially ordered dependency graph from the program, and only remove nodes which are terminal in the dependency graph. Figure~\ref{fig:sa_cp} shows the expression tree and dependency graph for a ShapeAssembly program. Additionally, for ShapeAssembly we apply this procedure iteratively, since each pruning operation updates the dependency graph.

\subsection{Code Grafting (CG)}
\label{subsec-cg}

\begin{figure}[t!]
    \centering
   \includegraphics[width=1.0\linewidth]{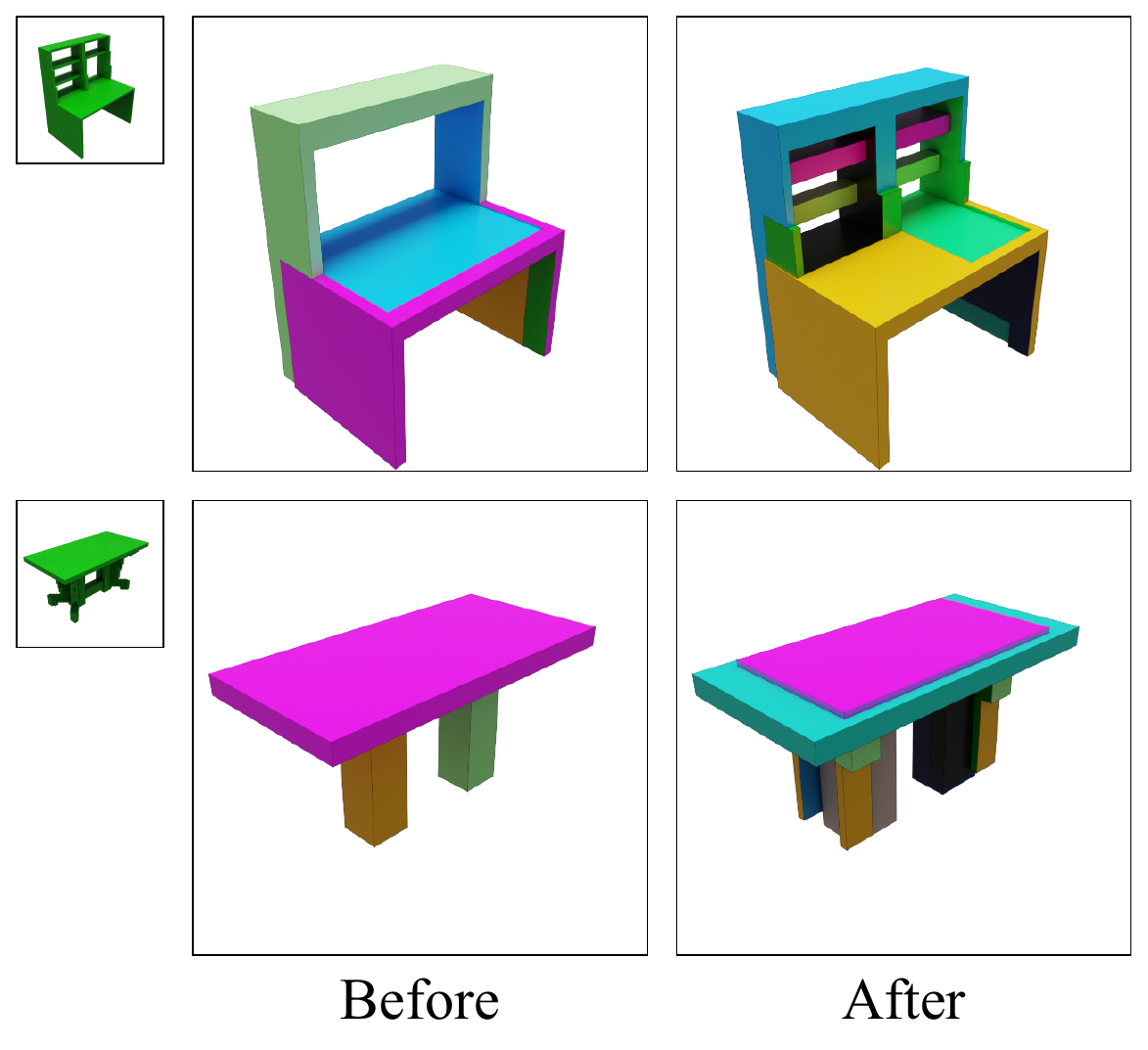}
    \caption{
    We show the \cgshort rewriter applied to 3D CSG programs.
    }
    \label{fig:cg_examples} 
    
\end{figure}

The \cgshort rewriter aims to replace sub-expressions of the given program with more suitable expressions from a cache of previously discovered sub-expressions.
Algorithm~\ref{alg-cg} provides a high-level overview of \cgshort, and Figure~\ref{fig:cg_examples} shows examples of applying this rewriter.

First, sub-expressions along with their executions (in the form of $n$-dimensional occupancy fields) are extracted from all the inferred programs. All the sub-expressions are then clustered by their execution using the FAISS~\cite{johnson2019billion_faiss} library (using \textsc{IndexBinaryIVF}) and stored in a cache. We store only \textit{unique} executions in the cache by identifying sub-expressions with approximately matching executions (hamming distance $< 100$ for 3D, and $<10$ for 2D) and retaining only the shortest `preferred' subexpression, while rejecting the rest. 

This step provides us our first rewrite. For each of the rejected sub-expressions, we retrieve the programs it originated from and insert the `preferred' subexpression into it. This rewrite results in programs which achieve similar reconstruction accuracy while having a shorter length. Lines $1$-$12$ in Algorithm~\ref{alg-cg} correspond to these steps.

The second rewrite performed in \cgname aims to replace sub-expressions in a given program to improve its reconstruction accuracy.

\begin{algorithm}[t]
\caption{\cgname}\label{alg-cg}
\begin{flushleft}
\textbf{Input:} Set of programs $Z_x$, sub-expression cache $C$ \& Executor $E$.\\
\textbf{Hyper-parameters:} $n$, $k=10$, $\tau=10$. \\
\textbf{Output:} Rewritten programs $Z_R$\\
\vspace{-0.5em}
\hrulefill
\vspace{-0.5em}
\begin{algorithmic}[1]
\STATE {\color{gray} \# creating the subexpression cache}
\STATE \textbf{for} $z_x$ in $Z_x$:
\STATE $\quad$ \textbf{for} \textit{subexpr} in extract\_subexprs($z_x$):
\STATE $\quad \quad$ \textit{match} = retrieve($C$, $E(\textit{subexpr})$)
\STATE $\quad \quad$ \textbf{if} \textit{match}:
\STATE $\quad \quad \quad$ $C$.remove(\textit{match})
\STATE $\quad \quad \quad$ \textit{shorter}, \textit{longer} = compare(\textit{subexpr}, \textit{match}) 
\STATE $\quad \quad \quad$ $C$.insert(\textit{shorter})
\STATE $\quad \quad \quad $ $z_r$ = replace($z_{longer}$, \textit{shorter})
\STATE $\quad \quad \quad$ $Z_R$.insert($z_r$)
\STATE $\quad \quad$ \textbf{else}:
\STATE $\quad \quad \quad$ $C$.insert(\textit{subexpr})
\STATE  {\color{gray} \# rewriting programs}
\STATE \textbf{for} $z_x$ in random\_sample($Z_x$, $n$):
\STATE $\quad$ num\_rewrites = 0 \& $z_r$ = $z_x$
\STATE $\quad$ \textbf{while}(num\_rewrites $< \tau$):
\STATE $\quad \quad$ \textit{candidates} = []
\STATE $\quad \quad$ \textbf{for} \textit{subexpr} in extract\_subexprs($z_r$):
\STATE $\quad \quad \quad$ $e^*$ = desired-execution(\textit{subexpr}, $E(z_x)$)
\STATE $\quad \quad \quad$ \textit{candidates}.extend($C$.get\_nn($e^*$, $k$))
\STATE $\quad \quad$ $z_{best}$ = get\_best(\textit{candidates})
\STATE $\quad \quad$ \textbf{if} $\mathcal{O}(z_{best}, x) \ge \mathcal{O}(z_{r}, x)$:
\STATE $\quad \quad \quad$ $z_r$ = $z_{best}$
\STATE $\quad \quad \quad$ num\_rewrites $\mathrel{+}= 1$
\STATE $\quad $ \textbf{if} $\mathcal{O}(z_r, x) \ge \mathcal{O}(z_{x}, x)$:
\STATE $\quad \quad$ $Z_{R}$.insert($z_{r}$)
\STATE \textbf{return} $Z_R$
\end{algorithmic}
\end{flushleft}
\end{algorithm}

We perform this rewrite in 3 steps:
\begin{packed_enumerate}
    \item We derive the \textit{desired execution} for each sub-expression in the program by masked function inversion (described in detail ahead). 
    Note that for each sub-expression we only consider the \textit{desired execution} within the bounding box of its execution (line 19).
    \item For each sub-expression we retrieve the $k$-nearest neighbors of its desired execution from the cache as its replacement candidates (line 20).
    \item We calculate the objective $\mathcal{O}$ achieved by each replacement, and perform the replacement which yields the highest reconstruction accuracy (lines 21-24).
\end{packed_enumerate}
This process (step 1 to 3) is repeated until none of the replacement candidates improve reconstruction accuracy, or until we perform a fixed number of replacements. Lines $14$-$24$ in Algorithm~\ref{alg-cg} correspond to these steps.

\begin{figure*}[t!]
    \centering
   \includegraphics[width=1.0\linewidth]{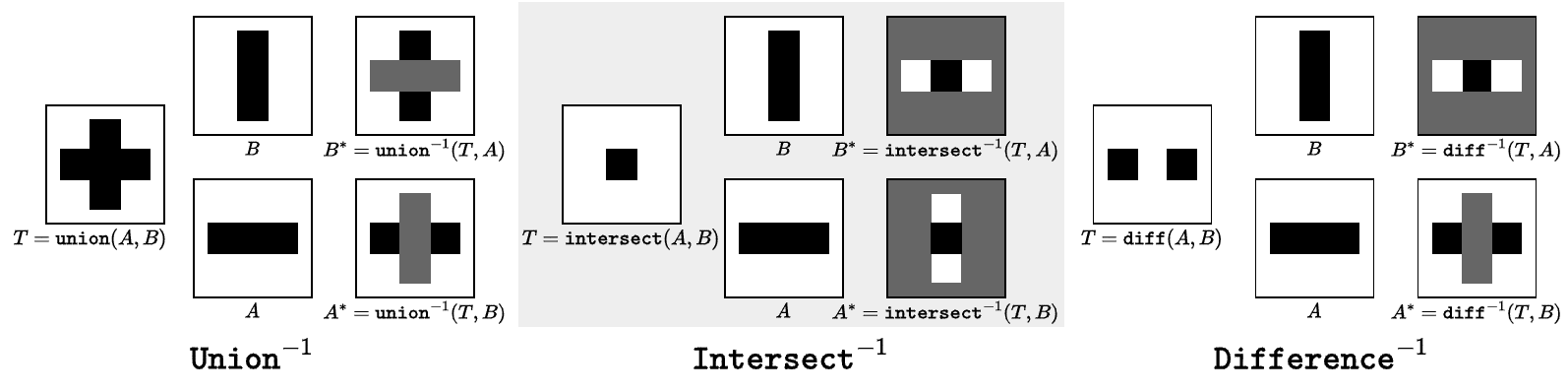}
    \caption{
    \cgshort~ rewriter derives \textit{desired executions} ($A^*$ and $B^*$)  for each sub-expression ($A$ and $B$) that can be used to search a cache for potential replacement candidates with respect to a target shape ($T$). The \textit{desired execution} is derived via masked function inversion, which we show the inversion for the three boolean combinators. For each desired execution, black indicates an area that should be occupied, white indicates an area that should not be occupied, and grey indicates invalid/masked regions.
    }
    \label{fig:op_inversion} 
\end{figure*}

\noindent
\textbf{Masked Function Inversion:} Given a sub-expression $A$, we derive its \textit{desired execution} $A^*$ by masked function inversion. 
We assume the given expression executes to the given target $T$, and invert the expression $S$. The process can be described as follows: 
\begin{align}
    T &\sim S(A) \\
    T &\sim S_1(S_2(... S_n(A))) \\
    A^* &\sim S^{-1}(T)\\
    A^*: &\sim S_n^{-1}(S_{n-1}^{-1}(...S_1^{-1}(T)))
\end{align}

By defining atomic inversion operations for transforms and combinators, we can simply derive $S^{-1}$ for a sub-expression $A$ by applying the inversions of operators applied on it, to the target $T$.
As $S$ can potentially be composed of non-invertible functions, we use a binary validity \textit{mask} 
to identify input regions of space ($\mathbb{R}^n$) that cannot be inverted 
, and perform inversion only for invertible regions.
For transforms such as translate, rotate, scale such masked inversions are trivial to define (e.g. $S_{translate}^{-1} = - S_{translate}$). For boolean combinators $U(A, B)$ we define its inversions w.r.t. a child $A$ as a $\{Target, Mask\}$ tuple as follows: 
\begin{align}
    \mathtt{Union}^{-1}(T, B) &= \{ T, \overline{\mathtt{Intersect}(T, B)}\}, \\
    \mathtt{Intersect}^{-1}(T, B) &= \{ T, \mathtt{Union}(T, B)\}, \\
    \mathtt{Diff}^{-1}(T, B) &= \{ T, \mathtt{Intersect}(T, \overline B)\}, \\
    \mathtt{Diff}^{-1}(T, A) &= \{ \tilde T, \mathtt{Union}(T, A)\},
\end{align}

where $\overline X$ stands for the complement of $X$, and the mask term indicates valid regions. We provide example of the inversions in Figure~\ref{fig:op_inversion}. 

\noindent
\textbf{Canonical Execution:} In CSG domain, all sub-expressions executions are used in \textit{canonical} form - we prepend each subexpression with a \textit{translate} and \textit{scale} command such that its execution is origin-centered and unit scale. 
Using the canonical form allows us to identify sub-expressions which are equivalent under translation/scale transformations. When the canonical sub-expressions are used for replacement, additional transform commands are prepended to make it fit the target expression's position and scale. For ShapeAssembly domain, all sub-programs are constructed in a unit scale cuboid by construction (their bounding box is fixed to sizes $(1, 1, 1)$). We note that similar canonical forms have been previously used in~\cite{2018-icfp-reincarnate} as well.

\noindent
\textbf{Empty Node:} During each \cgshort rewrite step, we optionally extend the input expression with a union and an empty node, i.e. $\textit{expr} = \text{Union}(\textit{expr}, \textit{empty})$. This allows us to additionally consider sub-expression from the cache which, when attached to the input expression, improve the objective.

\noindent
\textbf{Cache size:} We randomly sub-sample the cache as its size grows to curb the growth in memory requirement. We retain $35000$ subexpressions, each consisting $32 \times 32 \times 32$ binary values, resulting in only $\sim 1$ GB memory requirement. Note that the cache entries persist over multiple search-rewrite-train cycles, so that good sub-expressions are retained and propagated once they are discovered. 

\noindent
\textbf{ShapeAssembly:} We apply \cgshort to ShapeAssembly+, as it allows hierarchical composition of ShapeAssembly programs. Since the simple ShapeAssembly (used in the main draft) lacks hierarchical composition, we do not apply \cgshort to it.
First, we apply \cgshort to strictly replace or introduce ShapeAssembly sub-programs (rather than just a set of statements). This allows us to treat each ShapeAssembly program's implicit equivalent (as defined in Section~\ref{subsec-po}) as an simply-typed (partially invertible) lambda expression. To derive the \textit{desired executions}, we only need to invert simple CSG functions such as $\texttt{translate}$  and $\texttt{union}$. Successful expression replacement act as replacement (or introduction) of entire ShapeAssembly subprograms, mapping edited implicit equivalents uniquely to a ShapeAssembly program.

Note that the preconditions on the languages for applying \cgshort are fairly simple. It requires a mapping to a typed lambda calculus expression, which then allows us to perform type-matching replacements. Further, our masked inversion procedure requires the existence of masked inverse for each atomic operator.

\begin{figure*}[t]
    \centering
   \includegraphics[width=1.0\linewidth]{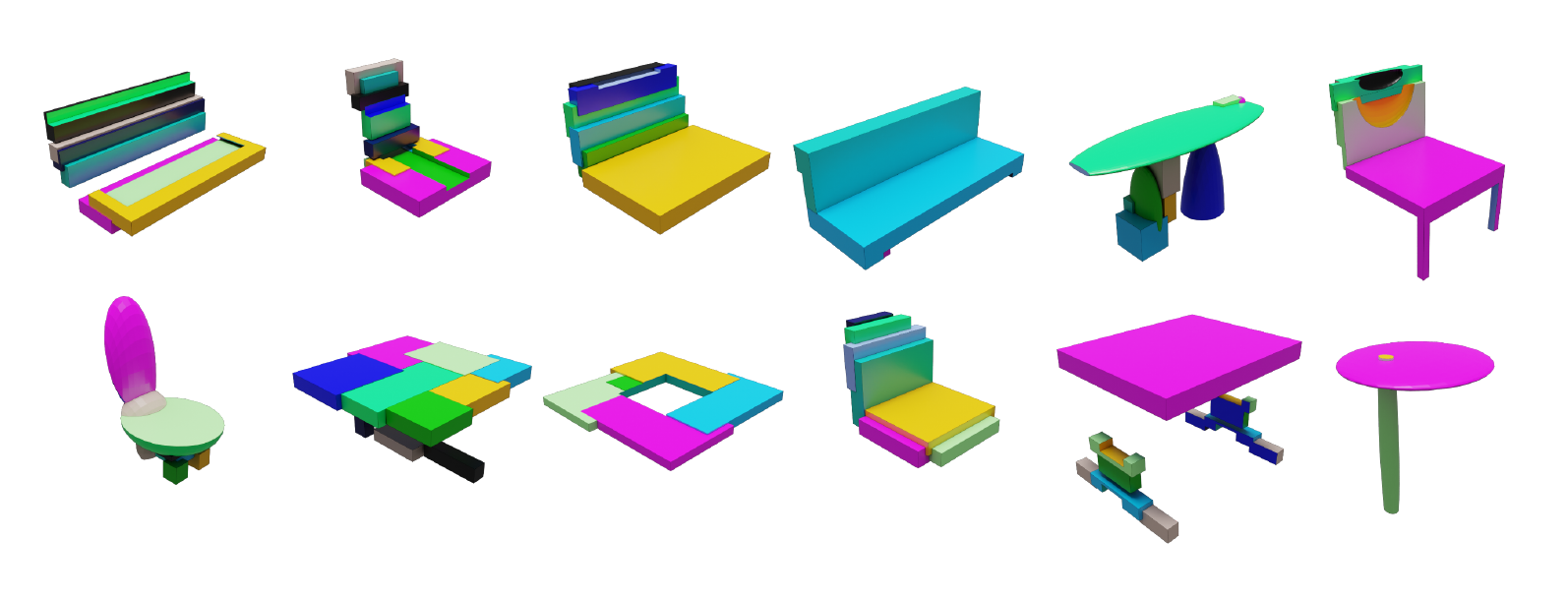}
    \caption{
    We show instances of program which fail to fully reconstruct the input shape, despite test time rewriting. A common trend we noticed is missing parts of object. Changing the rewrite objective $\mathcal{O}$ may help resolve this issue.
    }
    \label{fig:siri_failure} 
\end{figure*}

\begin{figure*}[t!]
    \centering
   \includegraphics[width=1.0\linewidth]{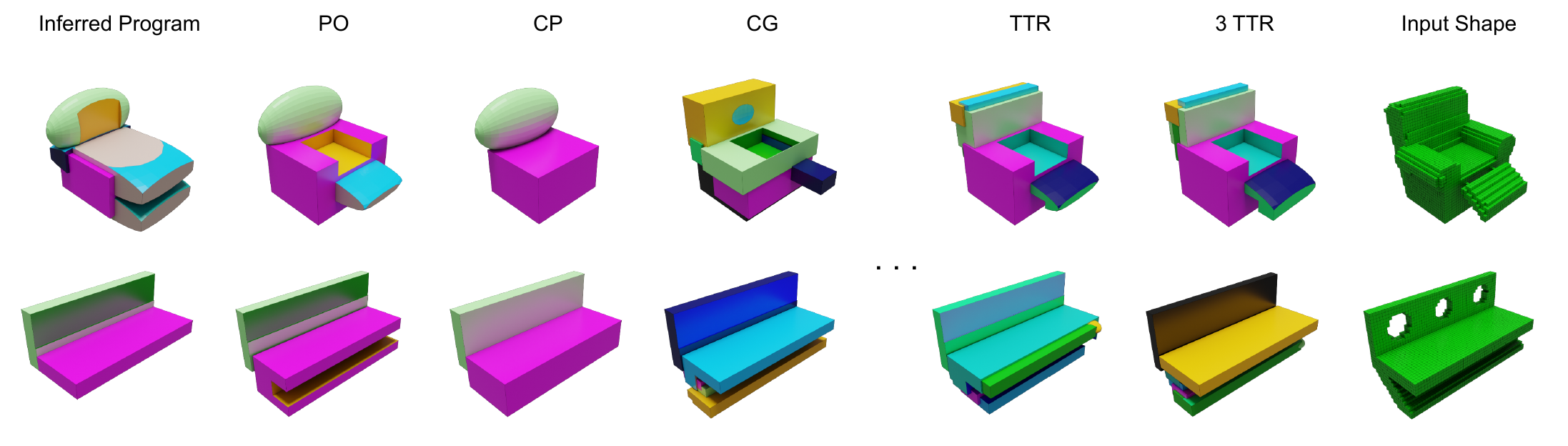}
    \caption{
    We show examples of the rewriter applied one at a time, and their interleaved application on inferred programs during test-time rewriting. Each rewriter is able to perform iterative improvements on the programs, and interleaved application further improves the program.
    }
    \label{fig:ttr_traj} 
\end{figure*}

\section{Qualitative Results}
\label{sec-qualitative}

We show qualitative comparisons between our method and prior approaches. Our method improves reconstruction accuracy over prior bootstrapping methods~\cite{jones2022PLAD}, and infers programs with higher conciseness than domain specific architectures~\cite{CSGSTUMP_ICCV}.  

\noindent
\textbf{Failure Cases:}  
While \methodshort~ achieves better aggregate reconstruction performance compared with previous boot-strapped learning methods such as PLAD, and matches reconstruction performance of domain-specific architectures such as CSG-Stump, its outputs can still be further improved.
One failure mode is that of missing parts, even after test-time rewriting (see Figure~\ref{fig:siri_failure}). 
We note \methodshort~ is not the only method that falls victim to this failure mode. 
A careful tuning of the program length weighting parameter $\alpha$ in the objective $\mathcal{O}$, along with a part presence sensitive reconstruction metric (instead of IoU) can help alleviate these challenges.
In fact, \methodshort~'s use of rewriters might allow it to uniquely solve this problem, by making use of a new class of rewriters that identifies missing semantic parts of a target shape, and rewrites the program with a sub-expression that covers these missing parts.

\noindent
\textbf{Test time rewriting:} We visualize test time rewriting in Figure~\ref{fig:ttr_traj}. As can be seen, each rewriter refines the program and their iterative application is beneficial.

\begin{figure*}[t!]
    \centering
   \includegraphics[width=0.96\linewidth]{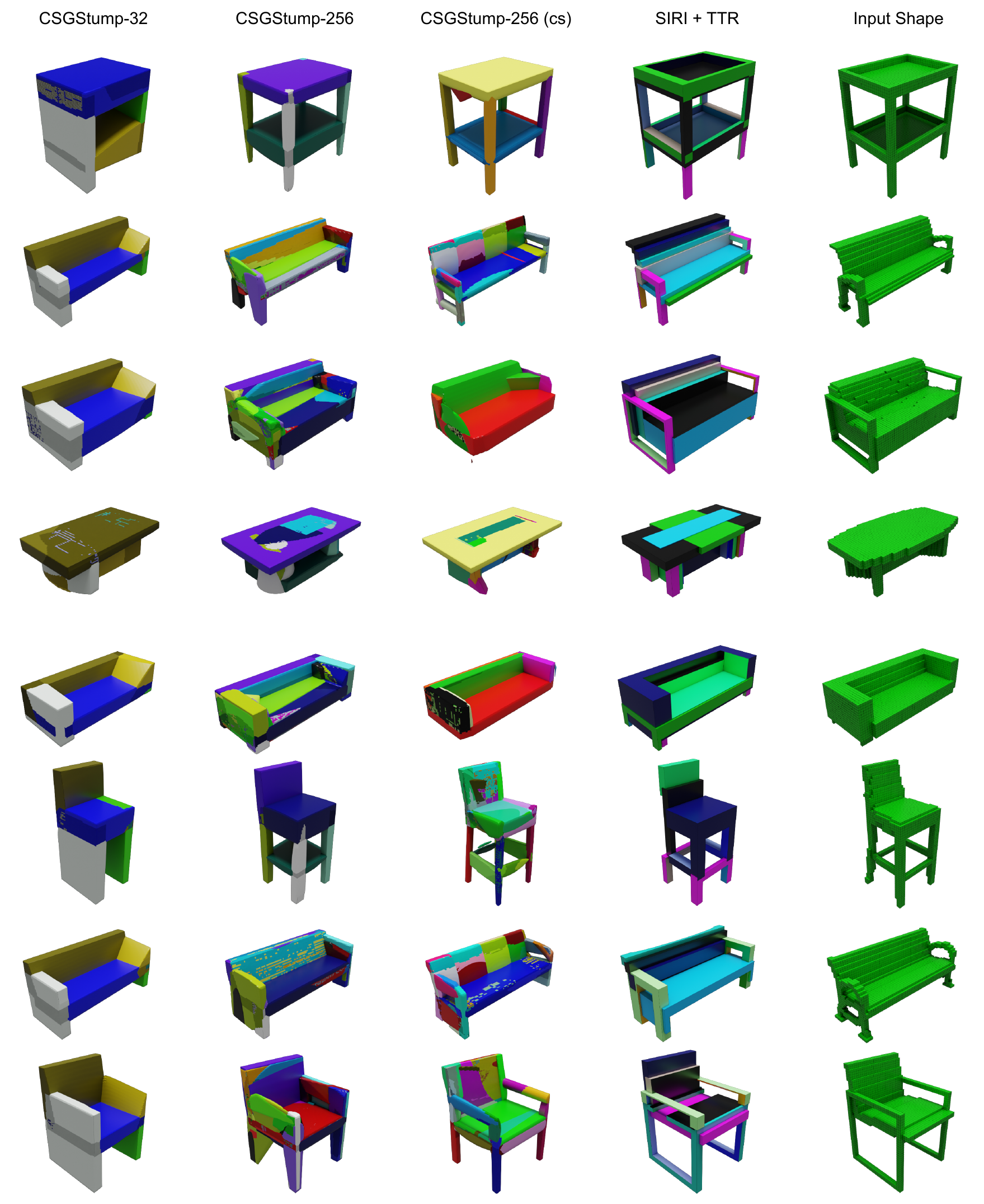}
    \caption{
   We compare our approach (\methodshort~ + test time rewrite) to three variants of CSGStump~\cite{CSGSTUMP_ICCV}: (a) \textit{CSGStump-32} a model trained with 32 primitive intersection nodes, (b) \textit{CSGStump-256} a model trained with 256 primitive and intersection nodes, and (c) \textit{CSGStump-256 (cs)} where a model is trained for each class (we use the pretrained weights provided by the authors). 
    }
    \label{fig:stump_comparison} 
\end{figure*}

\noindent
\textbf{Comparison to CSGStump:} We compare our method to three variants of CSG-Stump to \methodshort~ + TTR in Figure~\ref{fig:stump_comparison}. We note that while the class specific CSG-Stump model may surpass the reconstruction accuracy of \methodshort~ + TTR, its output predictions are still overly complex and hard to reason over, as reflected in the renderings with colored primitives. Note that due to the high complexity of CSG Stump programs, we color each intersection node, instead of the primitives. Despite this, we observe that the programs are still greatly over-parameterized.

\clearpage

\clearpage
{\small
\bibliographystyle{ieee_fullname}
\bibliography{egbib}
}